\documentclass{article}
\usepackage[dandb, final]{main}
\usepackage[dvipsnames, table]{xcolor}

\usepackage{graphicx}
\usepackage{amsmath}
\usepackage{amssymb}
\usepackage{booktabs}
\usepackage{mathtools}
\usepackage[accsupp]{axessibility}  
\usepackage{color}
\usepackage{array}
\usepackage{pdflscape}
\usepackage[longtable]{multirow}
\usepackage{longtable}
\usepackage{tabularray}
\usepackage{booktabs}
\usepackage{times}
\usepackage{epsfig}
\usepackage{amsmath}
\usepackage{bbm}
\DeclareMathAlphabet\mathbfcal{OMS}{cmsy}{b}{n}
\usepackage{makecell}
\usepackage{siunitx}
\usepackage{svg}
\usepackage{float}
\usepackage{comment}
\usepackage{lipsum}
\usepackage{stfloats}
\usepackage{multicol}
\usepackage{multirow}
\usepackage{bm}
\usepackage{etoolbox}
\usepackage{icomma}
\usepackage{array}
\usepackage{tabulary}
\usepackage{xcolor}
\usepackage{paralist}
\usepackage{booktabs}
\usepackage{adjustbox}
\usepackage{caption}
\usepackage{subcaption}
\usepackage{arydshln}
\usepackage{rotating}
\usepackage{enumitem}

\definecolor{gray}{rgb}{0.3,0.3,0.3}
\definecolor{blue}{rgb}{0,0.5,1}
\definecolor{mask_red}{rgb}{1,0,0.8}
\definecolor{green}{rgb}{0.2,1,0.2}
\definecolor{rblue}{rgb}{0,0,1}
\definecolor{lightblue}{HTML}{6495ed}
\definecolor{lightred}{HTML}{F19C99}
\definecolor{lightgreen}{rgb}{0.88, 1.0, 0.88}
\definecolor{lightyellow}{rgb}{1.0, 1.0, 0.85}

\definecolor{graytablerow}{gray}{0.6}

\usepackage{pifont}
\newcommand{\cmark}{\ding{51}}%
\newcommand{\xmark}{\ding{55}}%
\usepackage{tabu}
\usepackage[pro]{fontawesome5}
\usepackage{tikz}
\newcommand*\circled[1]{\tikz[baseline=(char.base)]{
\node[shape=circle,fill=gray,inner sep=0.5pt] (char) {\textcolor{white}{\small \textbf{#1}}};}}

\newcommand{\ie}{\emph{i.e.}}
\newcommand{\eg}{\emph{e.g.}}

\usepackage{wrapfig}

\newcommand*{\approxident}{%
  \mathrel{\vcenter{\offinterlineskip
  \hbox{$\sim$}\vskip-.35ex\hbox{$\sim$}\vskip-.35ex\hbox{$\sim$}}}}

\definecolor{lightblue}{rgb}{0.85, 0.93, 1.0}
\definecolor{lightgreen}{rgb}{0.88, 1.0, 0.88}
\definecolor{lightyellow}{rgb}{1.0, 1.0, 0.85}
\definecolor{cvprblue}{rgb}{0.21,0.49,0.74}

\usepackage[pagebackref,breaklinks,colorlinks]{hyperref}
\hypersetup{colorlinks, citecolor=cvprblue}

\usepackage[capitalize]{cleveref}
\crefname{section}{Sec.}{Secs.}
\Crefname{section}{Section}{Sections}
\Crefname{table}{Table}{Tables}
\crefname{table}{Tab.}{Tabs.}

\title{CHAOS: Chart Analysis with Outlier Samples}

\author{
\textbf{Omar Moured}$^{1,}$\thanks{Equal contribution. $^{\dagger}$Corresponding. } \qquad \textbf{Yufan Chen}$^{1,*}$ \qquad \textbf{Ruiping Liu}$^{1}$ \qquad \textbf{Simon Reiß}$^{1}$ \\ \textbf{Philip Torr}$^{2}$ \qquad \textbf{Jiaming Zhang}$^{1,\dagger}$ \qquad \textbf{Rainer Stiefelhagen}$^{1}$ \\
$^{1}$ Karlsruhe Institute of Technology \qquad $^{2}$ University of Oxford 
}

\begin{document}

\renewcommand\twocolumn[1][]{#1}%
\maketitle
\begin{center}
    \centering
    \captionsetup{type=figure}
    \includegraphics[width=\textwidth]{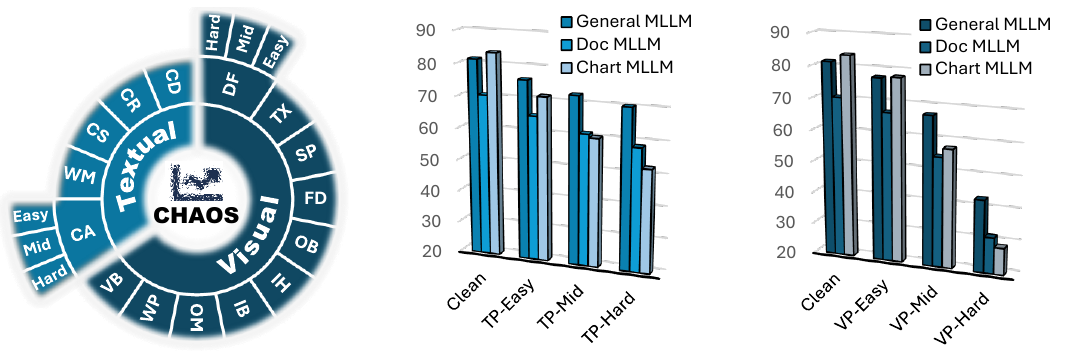}
    \begin{minipage}[t]{.4\textwidth}
        \vskip -3ex
        \subcaption{CHAOS benchmark. }\label{fig1a}
    \end{minipage}%
    \begin{minipage}[t]{.30\textwidth}
        \vskip -3ex
        \subcaption{Results on \textbf{TP}.}\label{fig1b}
    \end{minipage}%
    \begin{minipage}[t]{.30\textwidth}
        \vskip -3ex
        \subcaption{Results on \textbf{VP}.}\label{fig1c}
    \end{minipage}%
    \caption{(a) \faChartLine[regular] \textbf{CH}art \textbf{A}nalysis with \textbf{O}utlier \textbf{S}amples (\textbf{CHAOS}) benchmark includes 5 types of textural perturbations (\textbf{TP}) and 10 types of visual perturbations (\textbf{VP}), where each has 3 levels (\textit{Easy}, \textit{Mid}, \textit{Hard}). Results of general, document- and chart-specific MLLMs are compared on (b) textual perturbations and (c) visual perturbations with the relaxed accuracy (RA) scores.}
    \label{fig1:banner}
\end{center}%

\begin{abstract}
Charts play a critical role in data analysis and visualization, yet real-world applications often present charts with challenging or noisy features. However, ``outlier charts'' pose a substantial challenge even for Multimodal Large Language Models (MLLMs), which can struggle to interpret perturbed charts. In this work, we introduce \textbf{CHAOS (CHart Analysis with Outlier Samples)}, a robustness benchmark to systematically evaluate MLLMs against chart perturbations. CHAOS encompasses five types of textual and ten types of visual perturbations, each presented at three levels of severity (easy, mid, hard) inspired by the study result of human evaluation. The benchmark includes 13 state-of-the-art MLLMs divided into three groups (\ie, general-, document-, and chart-specific models) according to the training scope and data. Comprehensive analysis involves two downstream tasks (ChartQA and Chart-to-Text). Extensive experiments and case studies highlight critical insights into robustness of models across chart perturbations, aiming to guide future research in chart understanding domain. Data and code are publicly available at: \url{http://huggingface.co/datasets/omoured/CHAOS}.

\end{abstract}

\section{Introduction}
\label{sec:intro}

Much of humankind's knowledge is accessible through documents where information is condensed in structured visualizations. 
Through spending time reading and interpreting structured visuals, we as humans can gather insights about the content,~\eg, that the best performance in Fig.~\ref{fig1:banner} for a scenario \emph{TP-Hard} is a \emph{General-specific MLLM}.
Of course to structure different data, not only bar charts, but tables, line plots, scatter plots and general figures are used, which increases the complexity in processing them, even more so when doing it automatically through algorithmic means~\cite{kahou2017figureqa,methani2020plotqa,moured2023line}.
The emergence of Multimodal Large Language Models (MLLMs)~\cite{liu2024improved} has helped in the endeavour to interpret such structured chart data automatically~\cite{moured2024altchart,zhang2024tinychart,han2023chartllama}.
As such automatically answering textual questions about charts, so called chart question answering (ChartQA)~\cite{masry2022chartqa}, with MLLMs has seen steep improvement in recent years, yet, a major blind spot remains: \emph{How well can current MLLMs recover from corrupted chart data?}
This is a pressing question, as usage of multi-modal models in the real-world -- far away from clean testbeds --  increases, \eg, with visually impaired persons using models~\cite{jiang2024bench,moured2023accessible} as helping hand to understand physical documents.

In this work, we aim at shedding light into the darkness, by quantifying the susceptibility of chart question answering models towards real-world perturbations.
To achieve this, we design a comprehensive benchmark, the~\emph{Chart Analysis with Outlier Samples} (CHAOS) testbed (see Fig.~\ref{fig1a}), where we investigate the effects of ten visual perturbations (VPs) which are applied to images as well as five textual perturbations (TPs) that alter the textual inquiry.
With this, we can, for the first time get a hold of the effect that faulty camera sensors, badly lit scenes, speckles on the camera lens, typos, noisy speech recognition tools and many more errors have on current multi-modal chart interpretation models.
Furthermore, by rooting our benchmark in human perception through a user study, we are able to categorize the severity of perturbations into \textit{easy}, \textit{middle} and \textit{hard} tasks for humans and study how models perform along these difficulty levels. The performance of MLLMs and their degradation trends across severity levels are presented in Fig.~\ref{fig1b} for TPs and Fig.~\ref{fig1c} for VPs. More analysis of the results will be presented in the experiments.  

Furthermore, the proposed CHAOS benchmark includes two chart-related multimodal tasks, \ie, ChartQA and Chart2Text. To evaluate the robustness of MLLMs, we design a practical metric by considering both the original performance on clean chart data and the absolute drop when data is perturbed. Our benchmark involves 13 state-of-the-art MLLMs that focus on general-, document- and chart-specific tasks. As such, we provide critical insights into the robustness of MLLMs across visual and textual perturbations, aiming to guide future research in chart analysis. 

To summarize, our contributions are as follows: 
\begin{itemize}[leftmargin=1cm]
    \item A novel robustness benchmark for \textbf{CH}art \textbf{A}nalysis with \textbf{O}utlier \textbf{S}amples (\textbf{CHAOS}) is created
    . The multimodal benchmark includes 10 types of visual perturbations and 5 types of textual perturbations. Two tasks, \ie, chart summarization and chart QA, are included. 
    \item A pre-study human evaluation involving $42$ participants is conducted to finalize and construct the levels of severity for each visual perturbation. 
    \item A comprehensive analysis is performed by including 
    8 state-of-the-art MLLMs for general-, document- and chart-specific tasks. Through quantitative results and qualitative case study, there are different findings concluded for creating robust MLLMs. 
    \item A novel evaluation metric is proposed to perform robustness analysis according to relative and absolute performance degradation under various perturbations. 
    
\end{itemize}

\section{Related Work}
\label{related_work}

\noindent\textbf{Chart Analysis.} 
Interpreting information from charts has gained traction in the computer vision community, as it supports a variety of tasks aimed at understanding visual data. Information extraction~\cite{luo2021chartocr, mustafa2023charteye, liu2019data, ma2021towards} from charts involves detecting and decoding graphical elements to transform visual data into textual or numerical meta-data that are more accessible for further analysis. Besides, question-answering (QA) tasks~\cite{saleem2023realcqa, kantharaj2022opencqa, huang2024vprochart} related to charts enable systems to provide specific answers to user queries by deciphering the chart's data and layout, as exemplified by the ChartQA~\cite{masry2022chartqa} benchmark. Another critical task is summarization~\cite{liu2024chartthinker, krichene2024faithful, moured2024altchart}, where the system generates a concise textual summary that captures the main insights and trends depicted in the chart. Chart2Text~\cite{kantharaj2022chart} and CharSumm~\cite{rahman2022chartsumm} benchmarks evaluate systems' ability to generate textual summaries from charts, focusing on coherence and accuracy. These models often assume clean input data and lack systematic evaluation under real-world scenarios, which our work addresses by introducing the CHAOS benchmark for robustness assessment.

\noindent\textbf{Multimodal Large Language Models.}
Building on the momentum of foundational language models, Llama~\cite{touvron2023llama} and LLava~\cite{li2024llava} represent significant advancements in the field of VLMs, focusing on fine-tuning and visual instruction tuning to optimize performance across diverse language and vision tasks. These models are crucial for deeper multimodal learning, as evidenced by UNITER~\cite{chen2020uniter} and BLIP~\cite{li2022blip}, which refine how images and text interact. UNITER selects specific image areas while BLIP utilizes whole images. Specialized models like MatCha~\cite{liu2023matcha} and Pix2Struct~\cite{lee2023pix2struct} further tailor VLMs for specific content, focusing on contextual relevance in areas like chart understanding and screenshot parsing. This is completed by language integration into VLMs through GPT-3~\cite{floridi2020gpt} and CLIP~\cite{radford2021learningCLIP}, which employ natural language to enhance model adaptability and comprehension. 

\noindent\textbf{Robustness Benchmarks.}
Document restoration and rectification are related tasks aimed at enhancing the image quality of documents by correcting distortions. DocTr++~\cite{feng2023deep} explores unrestricted document image rectification. Research detailed in reference~\cite{fronteau2023evaluating} investigates robustness against adversarial attacks in document image classification. Auer~\textit{et al.}~\cite{auer2023icdar} introduced a challenge for robust document layout segmentation. In response, Zhang~\textit{et al.}~\cite{zhang2023welayoutwL} developed a WeChat layout analysis system. The robustness evaluation using the RVL-CDIP dataset~\cite{harley2015evaluation} focuses on document classification. Tran~\textit{et al.}~\cite{tran2017robust} designed a robust Document Layout Analysis (DLA) system leveraging a multilevel homogeneity structure. Chen \textit{et al.}~\cite{chen2024rodla} presented RoDLA, a robustness benchmark for Document Layout Analysis models, featuring a comprehensive taxonomy of document perturbations and evaluating models across various perturbations. Nevertheless, these efforts primarily concentrate on general document analysis and do not specifically address the unique challenges of chart interpretation under perturbations. Our CHAOS benchmark fills this gap, addressing both visual and textual perturbations specific to charts.

\section{CHAOS Benchmark}
\label{sec:method}
\begin{figure*}
    \centering
    \includegraphics[width=\linewidth]{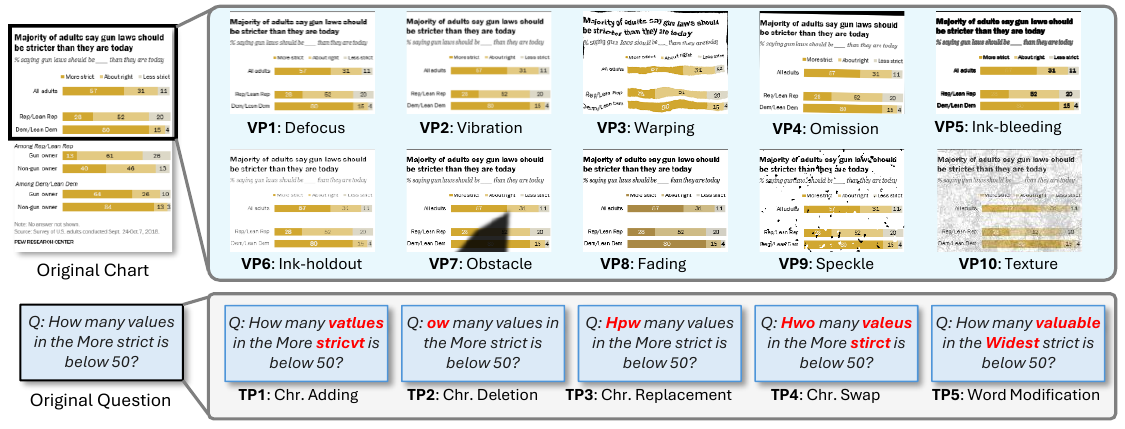}
    \caption{\textbf{Visualization of CHAOS benchmark} with 10 types of visual perturbations (VPs) and 5 types of textual perturbations (TPs).}
    \label{fig:pert_visual}
\end{figure*}

To assess the robustness of chart models, we introduce the CHAOS benchmark, encompassing a wide range of outlier samples reflecting common visual perturbations (VP) and textual perturbations (TP), as described in Table~\ref{tab:perturbation_list} and visualized in Fig.~\ref{fig:pert_visual}.

\begin{table}[ht]
  \centering
  \caption{\textbf{Perturbations Taxonomy} including Visual Perturbations (VPs) and Textual Perturbations (TPs) in CHAOS. Each perturbation has three difficulty levels (\textit{easy}, \textit{middle}, \textit{hard}).}
  \label{tab:perturbation_list}
  \resizebox{\textwidth}{!}{
    \setlength{\tabcolsep}{6.pt}
  \begin{tabular}{@{}llp{9.5cm}@{}}
    \toprule
    \textbf{ID} & \textbf{Perturbation Type} & \textbf{Perturbation Description} \\
    \midrule
    \multicolumn{3}{l}{\textbf{\textit{Visual Perturbation}}}\\
    VP1  & Defocus (DF)      & Convolve with a Gaussian kernel $G_{\sigma}$. \\
    VP2  & Vibration (VB)    & Apply a linear motion-blur kernel $K_{\text{motion}}$ (length $L$, angle $\theta$). \\
    VP3  & Warping (WP)      & Map pixels via a non-linear spatial transform $(x',y') = T(x,y)$. \\
    VP4  & Omission (OM)     & Random shifts $(\Delta x,\Delta y)$ and rotation $\theta$.\\
    VP5  & Ink-Bleeding (IB) & Morphological dilation expands dark regions. \\
    VP6  & Ink-Holdout (IH)  & Morphological erosion shrinks inked regions. \\
    VP7  & Obstacle (OB)     & Overlay an occlusion mask $O(x,y)$. \\
    VP8  & Fading (FD)       & Apply a linear transform $I' = \alpha I + \beta$ ($\alpha<1$) . \\
    VP9  & Speckle (SP)      & Add multiplicative noise $I' = I + I\! \cdot\! N$ with $N\!\sim\!\mathcal N(0,\sigma^{2})$. \\
    VP10 & Texture (TX)      & Blend with texture image $T$: $I' = \lambda I + (1-\lambda)T$. \\
    \midrule
     \multicolumn{3}{l}{\textbf{\textit{Textual Perturbation}}}\\
    TP1 & Character Adding (CA)      & Insert extraneous characters into words/sentences. \\
    TP2 & Character Deletion (CD)    & Randomly remove characters. \\
    TP3 & Character Replacement (CR) & Substitute correct characters with incorrect ones. \\
    TP4 & Character Swap (CS)        & Swap adjacent characters. \\
    TP5 & Word Modification (WM)     & Replace words with semantically nearby terms. \\
    \bottomrule
  \end{tabular}}
\end{table}

\subsection{Study Design}
To align CHAOS benchmark with human perception and practical scenarios, we establish 10 theoretical levels of perturbation based on parameters following~\cite{chen2024rodla, hendrycks2019robustness, hendrycks2021nae, imagenete}. To determine three meaningful difficulty levels, \ie, \textit{easy}, \textit{middle}, and \textit{hard}, we conducted an online user study involving 42 participants to complete a 10-question chart survey. All participants were presented with chart images subjected to different severity. For each perturbation, they were asked whether the chart is \textit{interpretable} and answer its question; if not, they proceeded to a less severe level (\eg, from L10 to L9). This process was repeated across all perturbations with unique chart images from ChartQA~\cite{masry2022chartqa} dataset. More details of study design are in supplementary~\ref{supply:human_eval_details}. This study enabled us to select three representative levels of severity that match up with real-world human experiences.

\subsection{Study Outcomes}

\begin{wrapfigure}[21]{r}{0.55\textwidth}
    \centering
    \includegraphics[width=0.99\linewidth]{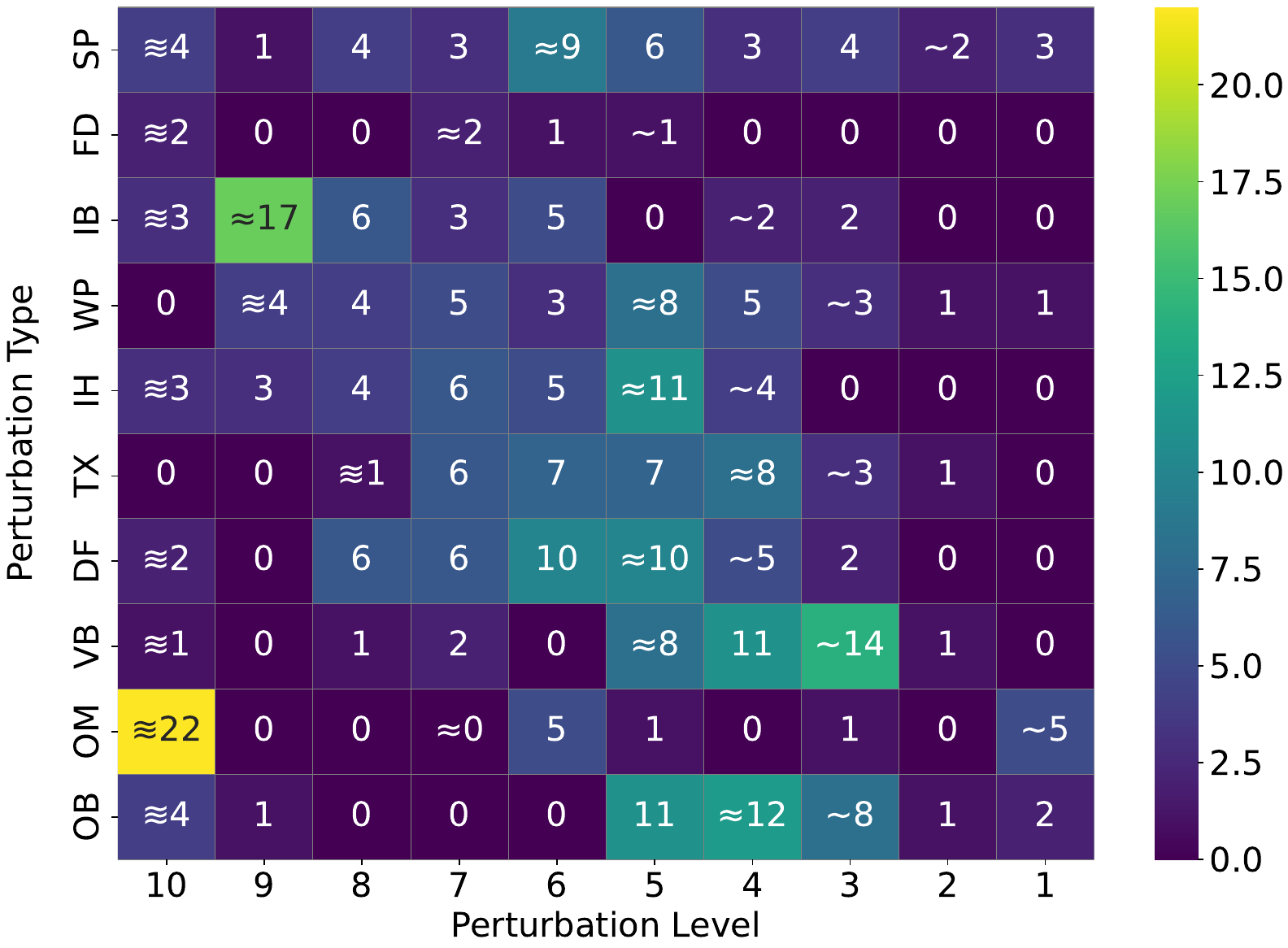}
    \caption{Distribution of 
    human study results across perturbation types (y-axis) and levels (x-axis). Each cell shows the number of participants who answered correctly. Symbols \{$\approxident$, $\approx$, $\sim$\} in the cell mean \{\textit{hard}, \textit{middle}, \textit{easy}\} levels for each perturbation (each row).
    }
    \label{fig:user_heatmap}
\end{wrapfigure}

The result of human evaluation, depicted in Fig.~\ref{fig:user_heatmap}, reveal a meticulous understanding of how humans perceive and interpret charts under varying levels of perturbation. While there is a general trend indicating that increased perturbation levels adversely affect interpretability, the extent of this impact differs decidedly across perturbation types. This suggests the necessity of customizing the definitions of \textit{easy}, \textit{middle}, and \textit{hard} levels for each perturbation.

For certain perturbations, correct responses were deficient across all levels. For instance, in the case of \textit{Fading} at higher severity, the charts appeared almost monochromatic. Participants often underestimated the loss of crucial color information, which was essential for tracing data points (additional examples are provided in the supplementary materials). On the other hand, some participants demonstrated innovative interpretive strategies with higher severity levels. In the presence of \textit{Speckle (SP)} noise, they estimated answers by focusing on chart elements less affected by noise, \eg, bars or lines with fewer speckles. This observation raises the question whether \textit{MLLMs can learn and incorporate these adaptive human  strategies to improve their robustness against severe perturbations?}

Based on the results from our human evaluation (Fig.~\ref{fig:user_heatmap}), we established the following three severity levels by using precise criteria that reflect human perceptual thresholds.

\noindent\textbf{(1) Easy Level} is defined as the highest level at which at least 90\% of participants could correctly interpret the chart and answer the associated question. Starting from Level 10, we incrementally increase the severity until this threshold is reached. This level represents conditions where perturbations have minimal impact on human interpretability. 

\noindent\textbf{(2) Middle Level} is determined by identifying the statistical mode of correct responses, \ie, the perturbation level at which the majority of participants succeeded. 
Unlike the mean, which can be skewed by outliers, the mode offers a stable central point where participants most commonly succeed, which captures a realistic balance between interpretability and perturbation effects. 

\noindent\textbf{(3) Hard Level} is defined as the highest perturbation level at which at least one participant was able to answer correctly, which represents the upper boundary of human interpretability. 

\noindent For instance, in the \textbf{Warping (WP)} of Fig.~\ref{fig:user_heatmap}, level 3 is designated as \textbf{easy}~(marked as ${\sim}$) since cumulatively from level 10 to level 3, at least 90\% of participants answered correctly. Level 5, where the number is the mode, is defined as \textbf{middle}~(marked as ${\approx}$). Level 9, with 4 participants answering correctly, is assigned as the \textbf{hard}~(marked as ${\approxident}$). These levels are designed to align the perturbations with realistic challenges that users might encounter.

\subsection{Metric of Robustness}\label{sec3.4:metric}

To fairly assess robustness in chart models, we propose a novel metric, $\mathcal{R}$, that considers both model performance degradation under perturbation and clean performance. Prior methods~\cite{hendrycks2019robustness, hendrycks2021nae, imagenete} tend to overlook the impacts of clean performance on robustness scores, which would misrepresent robustness by treating equal degradation in models with different performance as equivalent. Our proposed metric adjusts to offer a balanced assessment that aligns degradation with clean performance, as follows:  

\begin{equation}\label{eq1}
\mathcal{R}obustness = \frac{1}{X} \sum^{X}_{x=1} \left( 1 - \frac{1 - A_x}{\left(\frac{A_x}{A_{\text{clean}}}\right)^2 + \frac{1}{A_{\text{clean}}}} \right),
\end{equation}

\begin{wrapfigure}[24]{r}{0.55\textwidth}
  \centering
  \includegraphics[width=0.99\linewidth]{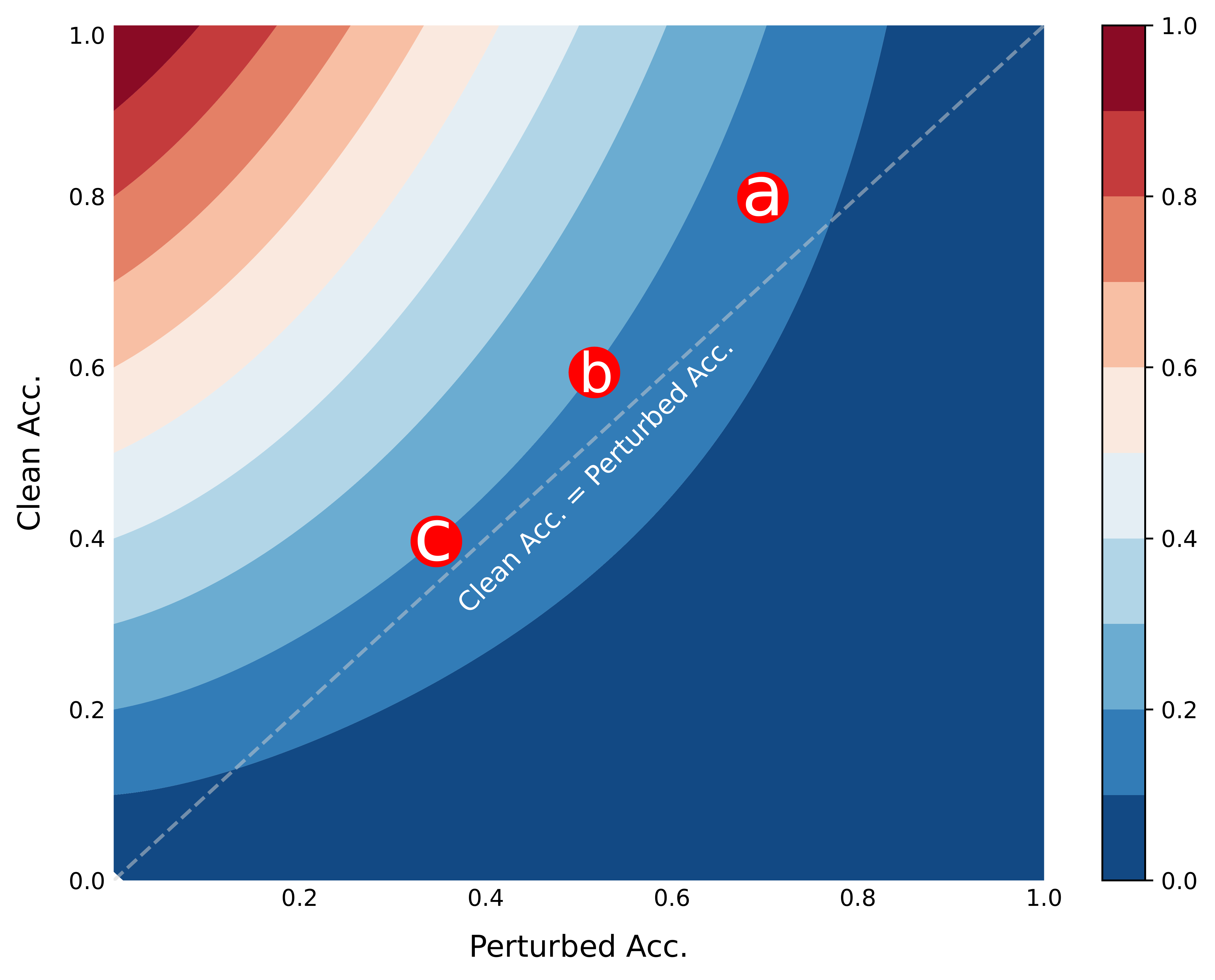}
  \vskip -0.5em
  \caption{Visualization of the metric ${\mathcal{R}}$ across perturbed and clean accuracy. All models on the same `contour' have the same ${\mathcal{R}}$ score. For the same absolute drop (clean${\rightarrow}$perturbed), the model with a lower clean accuracy has a lower robustness. \textit{E.g.}, $\mathcal{R}_{a}{>}\mathcal{R}_{b}{=}\mathcal{R}_{c}$, when $a{=}(0.7, 0.8), b{=}(0.5, 0.6), c{=}(0.33, 0.4)$. 
  }
  \vskip -1.5em
  \label{fig_r_metric}
\end{wrapfigure}

\noindent where $A_{\text{clean}}$ and $A_x$ represent the model's performance on clean and perturbed datasets, respectively, with $x$ indicating the perturbation level (\eg, easy, middle, hard). The ratio $\frac{A_x}{A_{\text{clean}}}$ captures the relative differential in performance and normalizes the perturbed accuracy, thus providing a proportional metric that is independent of the absolute performance magnitude. This adjustment ensures that the robustness metric reflects relative degradation, placing more splendid emphasis on models whose performance on clean data is high yet still experiences considerable drops under perturbation. Additionally, our metric incorporates the absolute performance degradation term $1 - A_x$, which emphasizes cases where the model's performance drops significantly. By combining relative and absolute degradation measures, our robustness score balances both dimensions of robustness assessment: it penalizes models that fail under perturbation more severely while rewarding those that can sustain performance even in challenging conditions. As shown in Fig.~\ref{fig_r_metric}, with three perturbation levels in our benchmark, the maximum possible robustness score is 1.0, achieved when no degradation occurs ($A_x = A_{\text{clean}}$), representing perfect robustness. Conversely, the minimum score is 0, which equals complete failure across all levels ($A_x = 0$ for all perturbations). Higher values of $\mathcal{R}$ indicate higher robustness, with the score furnishing a precise indication of the model’s robustness across perturbation levels.

\section{Experiments}
To benchmark CHAOS, we conducted experiments with 8 large VLMs across 2 distinct chart understanding tasks. Below, we outline the experimental setup (Sec.~\ref{sec4.1:implementation}), the main results and findings (Sec.~\ref{sec4.2:results}), the hallucination analysis (Sec.~\ref{sec4.3:hallucination}), and limitations (Sec.~\ref{sec4.4:limitations}). 

\subsection{Implementation Details}\label{sec4.1:implementation}

\subsubsection{Datasets} 

\noindent\textbf{ChartQA.} To analyze vision-language models in the CHAOS benchmark setup with both VPs and TPs, we utilize the ChartQA~\cite{masry2022chartqa} dataset, which includes 2.5K machine-augmented and human-annotated question-answer pairs. 
ChartQA features a diverse distribution of questions, including data retrieval, visual reasoning, compositional reasoning, and visual-and-compositional reasoning. 

\noindent\textbf{Chart-to-Text.} Involving different tasks on the CHAOS benchmark with TPs, we utilize the Chart-to-Text \cite{kantharaj2022chart} dataset, designed to generate captions summarizing key insights from a given chart. This dataset includes two real-world sources: Pew and Statista, covering a broad range of topics and five chart types. It comes with 6.61K test images.

\subsubsection{Evaluation Metric} 
For the ChartQA task, we utilize the Relaxed Accuracy (RA) metric, following \cite{methani2020plotqa,masry2022chartqa}. This metric accommodates minor inaccuracies in numerical value predictions, allowing a deviation of up to 5\% from the gold-standard answer. For nonnumerical answers, however, the predictions must exactly match the gold-standard answer to be considered correct. RA has subsequently become the standard metric for evaluating numerical answers.
For the Chart-to-Text task, we adopt BLEU-4 and Content Selection as the evaluation metrics, following \cite{kantharaj2022chart}.
For robustness, we compile the results across all perturbations and levels to compute the proposed robustness metric $\mathcal{R}$ in Eq.~(\ref{eq1}), including visual $\mathcal{R}_{VP}$ and textural $\mathcal{R}_{TP}$. It offers a holistic evaluation by incorporating both relative and absolute performance degradations.

\subsubsection{MLLM Baselines}
We categorize the models by training scope and data into three groups: \textit{general}, \textit{document}- and \textit{chart-related} MLLMs. Details of selecting MLLMs, please refer to the supplementary. 

\begin{itemize}[leftmargin=1cm]
\item General MLLMs: \textbf{LLaVA-OneVision}~\cite{li2024llava}, \textbf{InternVL2}~\cite{wang2024internvl2}, \textbf{GPT-4o}~\cite{hurst2024gpt4o}, \textbf{Qwen-VL}~\cite{bai2023qwen}, \textbf{Janus-Pro}~\cite{chen2025januspro} are pre-trained by using general vision-language data, such as image captioning, visual question answering, and image generation. 

\item Document-related MLLMs: \textbf{UReader}~\cite{ye2023ureader}, \textbf{DocOwl1.5}~\cite{hu2024mplugdocowl} and \textbf{DocOwl2}~\cite{hu2024mplugdocowl2} are more inclined to document analysis tasks, as they both are trained from document-related data to achieve a variety of document understanding tasks. 

\item Chart-related: \textbf{ChartInstruct}~\cite{masry2024chartinstruct}, \textbf{ChartLlma}~\cite{han2023chartllama}, \textbf{ChartAssistant}~\cite{meng2024chartassisstant}, \textbf{TinyChart}~\cite{zhang2024tinychart}, and \textbf{ChartMoE}~\cite{xu2024chartmoe} are fine-tuned on the downstream chart datasets like ChartQA and Chart-to-Text, specifically for chart understanding tasks. 
\end{itemize}

\subsection{Results on CHAOS Benchmark}\label{sec4.2:results}

\subsubsection{Results of ChartQA}

\begin{table*}[ht]
\caption{Results on CHAOS benchmark of ChartQA. \textbf{VP}: Visual Perturbations; \textbf{TP}: Textural Perturbations. The metrics include the relaxed accuracy (RA${\uparrow}$) for clean and three levels, the robustness ($\mathcal{R}{\uparrow}$). The absolute drops relative to clean RA are marked in \textcolor{red}{red}. }
\vspace{-8pt}
\label{tab:sota_chartqa}
\renewcommand{\arraystretch}{1.2}
\resizebox{\textwidth}{!}{
\setlength{\tabcolsep}{2.pt}

\begin{tabular}{@{}lrccc|c|ccc|c|ccc|c|c@{}}
\toprule
\multirow{2}{*}{Model} & \multirow{2}{*}{Year} & \multirow{2}{*}{\#Param} & \multirow{2}{*}{Resolution} & \multirow{2}{*}{\begin{tabular}[c]{@{}c@{}}Inference\\ Throughput\end{tabular}} & ChartQA & \multicolumn{3}{c|}{\textbf{CHAOS-VP}} & \multirow{2}{*}{$\mathcal{R}_{VP}$} & \multicolumn{3}{c|}{\textbf{CHAOS-TP}} & \multirow{2}{*}{$\mathcal{R}_{TP}$} & \multirow{2}{*}{$\mathcal{R} \uparrow $} \\ 
 &  &  &  & & Clean & Easy & Mid & Hard &  & Easy & Mid & Hard &  &  \\ \midrule

\rowcolor[gray]{.9} \multicolumn{2}{l}{\textbf{\emph{General}}} & & &  &  &  &  &  & &  &  & & - & \\

LLaVA-OneVision~\cite{li2024llava} & 2024 & 7B & 384\(\times\)384 & 1.27 it/s &  81.32 & 77.42 \textcolor{red}{(-3.90)} & 67.20 \textcolor{red}{(-14.12)} & 42.83 \textcolor{red}{(-38.49)} & 78.12 & 75.98 \textcolor{red}{(-5.34)} & 72.46 \textcolor{red}{(-8.86)} & 70.22 \textcolor{red}{(-11.10)} & 86.63 & 82.37 \\

InternVL2~\cite{wang2024internvl2} & 2024 & 8B & 448×448×Ada.* & 3.40 it/s & 85.08 & 80.99 \textcolor{red}{(-4.09)} & 67.83 \textcolor{red}{(-17.25)} & 38.68 \textcolor{red}{(-46.40)} & 76.24 & 78.18 \textcolor{red}{(-6.90)} & 72.53 \textcolor{red}{(-12.55)} & 69.10 \textcolor{red}{(-15.98)} & 85.97 & 81.11 \\ 

GPT-4o~\cite{hurst2024gpt4o} & 2024 & - & ** & 1.20 it/s & {72.48} & 69.88 \textcolor{red}{(-2.60)} & 62.39 \textcolor{red}{(-10.09)} & 45.51 \textcolor{red}{(-26.97)} & 79.50 & 66.60 \textcolor{red}{(-5.88)} & 62.86 \textcolor{red}{(-9.62)} & 61.43 \textcolor{red}{(-11.05)} & 83.06 & 81.28 \\

Qwen2.5-VL~\cite{Qwen2.5-VL} & 2025 & 7B & native resolution & 1.61 it/s & \textbf{87.84} & 85.51 \textcolor{red}{(-2.33)} & 75.24 \textcolor{red}{(-12.60)} & 49.89 \textcolor{red}{(-37.95)} & \textbf{81.84} & 81.97 \textcolor{red}{(-5.87)} & 77.18 \textcolor{red}{(-10.66)} & 75.30 \textcolor{red}{(-12.54)} & \textbf{88.63} & \textbf{85.24} \\

Janus-Pro~\cite{chen2025januspro} & 2025 & 7B & native resolution & 1.03 it/s & 60.04 & 50.33 \textcolor{red}{(-9.71)} & 38.90 \textcolor{red}{(-21.14)} & 25.62 \textcolor{red}{(-34.42)} & 69.82 & 52.26 \textcolor{red}{(-7.78)} & 46.98 \textcolor{red}{(-13.06)} & 43.14 \textcolor{red}{(-16.90)} & 76.99 & 73.41 \\
  
\midrule

\rowcolor[gray]{.9} \multicolumn{2}{l}{\textbf{\emph{Document-related}}} & &  &  &  &  &  &  & &  &  & & - & \\

DocOwl1.5~\cite{hu2024mplugdocowl} & 2024 & 8B & 448\(\times\)448(\(\times\)9) & 1.56 it/s & 70.50 & 66.98 \textcolor{red}{(-3.52)} & 54.69 \textcolor{red}{(-15.81)} & 31.37 \textcolor{red}{(-39.13)} & 73.63 & 65.24 \textcolor{red}{(-5.26)} & 61.12 \textcolor{red}{(-9.38)} & 58.46 \textcolor{red}{(-12.04)} & 82.36 & 77.99 \\ 

UReader~\cite{ye2023ureader} & 2023 & 7B & 224\(\times\)224(\(\times\)20) & 1.67 it/s & 59.30 & 52.88 \textcolor{red}{(-6.42)} & 42.19 \textcolor{red}{(-17.11)} & 26.30 \textcolor{red}{(-33.00)} & 71.84 & 54.32 \textcolor{red}{(-4.98)} & 49.54 \textcolor{red}{(-9.76)} & 46.85 \textcolor{red}{(-12.45)} & 79.25 & 75.54 \\

DocOwl2~\cite{hu2024mplugdocowl2} & 2024 & 8B & 448\(\times\)448(\(\times\)9) & 1.7 it/s & {69.68} & 66.77 \textcolor{red}{(-2.91)} & 53.33 \textcolor{red}{(-16.35)} & 29.68 \textcolor{red}{(-40.00)} & 73.10 & 64.30 \textcolor{red}{(-5.38)} & 60.02 \textcolor{red}{(-9.66)} & 57.78 \textcolor{red}{(-11.90)} & 82.04 & 77.57 \\ 

\midrule

\rowcolor[gray]{.9} \multicolumn{2}{l}{\textbf{\emph{Chart-related}}} & & & & &  &  &  & &  &  & & - & \\

ChartInstruct~\cite{masry2024chartinstruct} & 2024 & 7B & 512\(\times\)512 & 1.40 it/s &  66.64 & 38.35 \textcolor{red}{(-28.29)} & 27.37 \textcolor{red}{(-39.27)} & 16.64 \textcolor{red}{(-50.00)} & 56.50 & 40.56 \textcolor{red}{(-26.08)} & 34.54 \textcolor{red}{(-32.10)} & 30.50 \textcolor{red}{(-36.14)} & 63.53 &  60.02 \\

ChartLlama~\cite{han2023chartllama} & 2023 & 13B & 336\(\times\)336 & 1.94 it/s & 75.28 & 45.53 \textcolor{red}{(-29.75)} & 35.64 \textcolor{red}{(-39.64)} & 30.02 \textcolor{red}{(-45.26)} & 59.78 & 61.18 \textcolor{red}{(-14.10)} & 55.50 \textcolor{red}{(-19.78)} & 52.34 \textcolor{red}{(-22.94)} & 76.80 & 68.29
\\

ChartAst~\cite{meng2024chartassisstant} & 2024 & 13B & 448\(\times\)448 & 1.47 it/s & 79.90 & 48.28 \textcolor{red}{(-31.62)} & 37.96 \textcolor{red}{(-41.94)} & 24.94 \textcolor{red}{(-54.96)} & 56.79 & 50.77 \textcolor{red}{(-29.13)} & 45.49 \textcolor{red}{(-34.41)} & 42.80 \textcolor{red}{(-37.10)} & 66.16 & 61.48 \\

TinyChart@768~\cite{zhang2024tinychart} & 2024 & 3B & 768\(\times\)768 & 3.14 it/s & 83.60 & 77.88 \textcolor{red}{(-5.72)} & 57.45 \textcolor{red}{(-26.15)} & 28.47 \textcolor{red}{(-55.13)} & 69.76 & 71.37 \textcolor{red}{(-12.23)} & 60.10 \textcolor{red}{(-23.50)} & 52.27 \textcolor{red}{(-31.33)} & 77.25 & 73.50 \\ 

ChartMOE+PoT~\cite{xu2024chartmoe} & 2024 & 8B & 490\(\times\)490 & 1.44 it/s & 84.52 & 78.50 \textcolor{red}{(-6.02)} & 63.37 \textcolor{red}{(-21.15)} & 38.89 \textcolor{red}{(-45.63)} & 74.90 & 78.03 \textcolor{red}{(-6.49)} & 72.10 \textcolor{red}{(-12.42)} & 69.06 \textcolor{red}{(-15.46)} & 85.96 & 80.43 \\ 

\bottomrule

\end{tabular}

}
\vskip -2ex
\end{table*}

\noindent\textbf{{\faSearch} Finding 1: MLLM models are highly sensitive to minor pixel distortions.} In Table~\ref{tab:sota_chartqa}, we observe that even under easy perturbations, often nearly imperceptible to human observers, performance drops by at least 4\%. This substantially highlights their vulnerability to subtle pixel-level changes, which is frequently encountered in the real-world.

\noindent\textbf{{\faSearch} Finding 2: Robustness trade-off emerges from general to expert models.}
While general models may perform fewer task-specific understanding tasks, they exhibit the highest average robustness ($\overline{\mathcal{R}}_{Gen}=80.68$), followed by document-expert models ($\overline{\mathcal{R}}_{Doc}=77.03$). In contrast, chart-specialized models show lower average robustness ($\overline{\mathcal{R}}_{Chart}=68.7$). This discrepancy is further highlighted by the significant performance drops of chart-related models across perturbations, averaging 23.25 \% for the easy level and 50\% for medium and hard levels under visual perturbations.

\begin{wrapfigure}[16]{r}{0.6\textwidth}
  \centering
  \vskip -1.5em
  \includegraphics[width=0.99\linewidth]{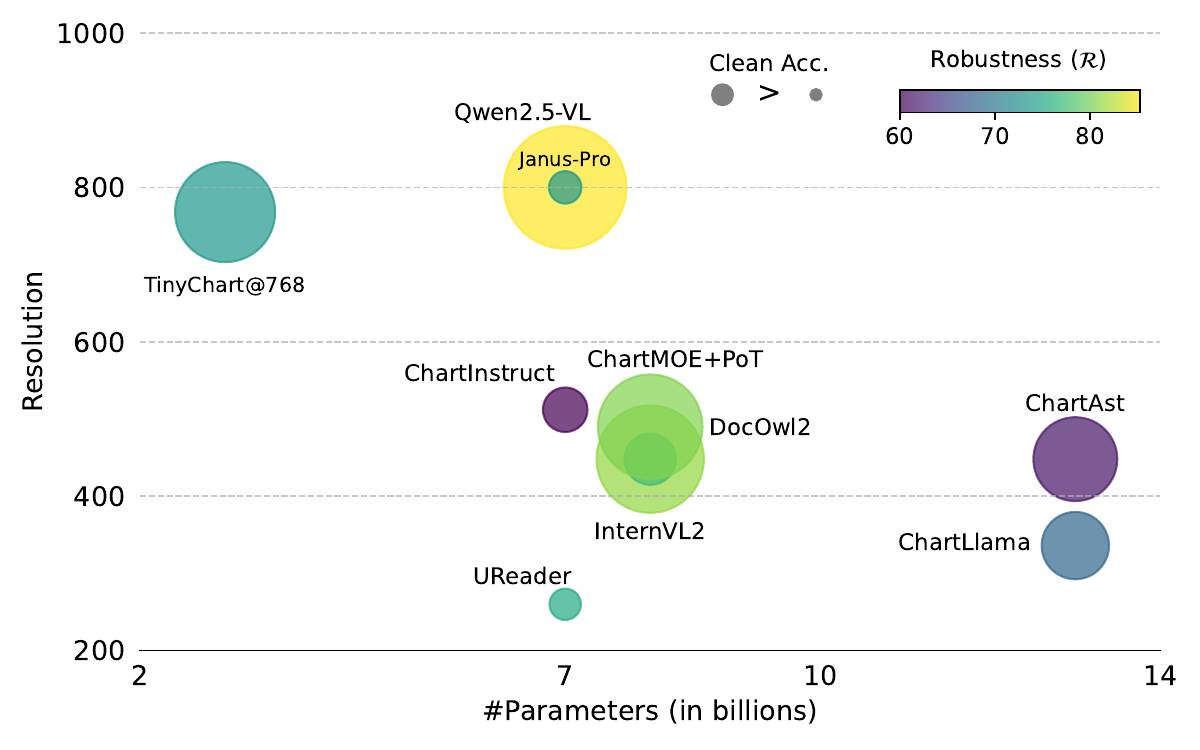}
  \vskip -1em
  \caption{Robustness analysis. The clean accuracy is represented by the circle size, while robustness is by color intensity, with lighter colors for higher robustness. }
  \label{fig_findingsupport}
\end{wrapfigure}

\noindent\textbf{{\faSearch} Finding 3: Textual perturbations can be just as significant as visual perturbations.} A closer examination of VP and TP reveals that textual distortions can be as impactful as visual ones. Clean inputs do not inherently guarantee consistent performance, as TP alone can cause a significant 31\% performance drop. Such results emphasize the often-overlooked importance of textual distortions.

\noindent\textbf{{\faSearch} Finding 4: Robustness is a result of multiple factors.} As shown in Fig.~\ref{fig_findingsupport}, despite chart-related models using higher-resolution inputs and larger parameter counts, they did not show better robustness. This suggests that these parameters alone are not sufficient to be robust. Further attention should be devoted to training data and fine-tuning strategy. While this conclusion holds within our benchmark, it needs further investigation.

\subsubsection{Results of Chart Summarization}
\begin{table*}[ht]
\caption{Results on CHAOS benchmark of Chart-to-Text. \textbf{VP}: Visual Perturbations. BLEU-4 and Content Selection as the evaluation metrics are reported for clean data and three VP levels. * The average inference time on perturbation is reported.
}
\vspace{-6pt}
\label{tab:sota_chart2text}
\resizebox{\textwidth}{!}{
\setlength{\tabcolsep}{12.0pt}

\begin{tabular}{@{}lrcc|c|ccc@{}}
\toprule
\multirow{2}{*}{Model} & \multirow{2}{*}{\#Param} & \multirow{2}{*}{Resolution} & \multirow{2}{*}{\begin{tabular}[c]{@{}c@{}}Inference\\ Time (in seconds)\end{tabular}} & Chart-to-Text & \multicolumn{3}{c}{\textbf{CHAOS-VP}} \\ 
\cmidrule(l){5-8}
 &  &  &  & Clean & Easy & Mid & Hard \\ 
\midrule

ChartInstruct~\cite{masry2024chartinstruct} & 7B & 512\(\times\)512 & 8.4\textsuperscript{*} & 13.83 & 5.16 \textcolor{red}{(-8.67)} & 3.89 \textcolor{red}{(-9.94)} & 2.30 \textcolor{red}{(-11.53)} \\

ChartLlama~\cite{han2023chartllama} & 13B & 336\(\times\)336 & 19.4\textsuperscript{*} & 14.23 & 3.30 \textcolor{red}{(-10.93)} & 2.81 \textcolor{red}{(-11.42)} & 1.89 \textcolor{red}{(-12.34)} \\

TinyChart@768~\cite{zhang2024tinychart} & 3B & 768\(\times\)768 & 25.12\textsuperscript{*} & 17.18 & 15.16 \textcolor{red}{(-2.02)} & 10.63 \textcolor{red}{(-6.55)} & 4.95 \textcolor{red}{(-12.23)} \\ 

\bottomrule
\end{tabular}
}
\vskip -2ex
\end{table*}

Summarization demands identifying key visual trends and articulating them into concise, coherent text. Table \ref{tab:sota_chart2text} shows the significant performance degradation on different chart task under perturbations, with an average drop of 7.21\% on easy levels, over 10\% on medium and hard. Analysis of over 50 cases reveals false factual hallucinations, repetitive phrases, and nonsensical outputs like "\textit{figure1.png}", highlighting the models' struggle with maintaining logical flow. Unlike QA tasks, which typically require single-word responses, chart-to-text tasks demand the recursive construction of longer token sequences. Our results highlight a tenfold increase in inference time, driven by the encoders' struggle to mitigate the effects of distortions. This underscores the significant impact of low robust MLLM.

\subsection{Hallucination Analysis}\label{sec4.3:hallucination}
Most existing MLLMs~\cite{bai2023qwen, liu2023improvedllava, liu2024llava, liu2023llava1.5, ye2023ureader, hu2024mplugdocowl} exhibit hallucination issues, such as predicting objects or content that does not exist in the input.

\noindent\textbf{Numerical Reasoning.} Approximately 40\% of chart-related questions, particularly from the human-authored split of ChartQA, involve reasoning tasks. These include visual, compositional, and visual-compositional reasoning, which require mathematical operations such as summation, subtraction, multiplication, and comparative analysis (e.g., determining higher, lower, or equal values). Models like TinyChart, which employ specialized techniques such as Program-of-Thoughts (PoT), and LLaVA-OneVision, explicitly trained on large-scale mathematical reasoning instructions, demonstrate significantly better performance on these tasks. In contrast, UReader which was primarily trained for text-reading tasks, achieves the lowest accuracy of 39.28\% on the human split.

\begin{wraptable}[16]{r}{0.5\textwidth}
    \centering
    \caption{Hallucination analysis with blank input images (completely black) on ChartQA. Relaxed Accuracy is reported.}
    \label{tab:blank_input_results}
    \resizebox{0.5\textwidth}{!}{%
        \begin{tabular}{lccc}
            \toprule
            \multirow{2}{*}{\textbf{Model}} & \multirow{2}{*}{\textbf{Image}} & \multicolumn{2}{c}{\textbf{ChartQA Official Split}} \\
            \cmidrule(lr){3-4}
            & & \textbf{Augmented} & \textbf{Human} \\
            \midrule
            Qwen2.5-VL      & \cmark & 94.96 & 80.72 \\
            LLaVA-OneVision & \cmark & 92.80 & 69.84 \\
            TinyChart       & \cmark & 94.80 & 57.92 \\ 	
            ChartMOE+PoT       & \cmark & 90.96 & 78.08  \\
            \midrule
            Qwen2.5-VL      & \xmark & 9.28 & 14.88 \\
            LLaVA-OneVision & \xmark & 13.76 & 15.68 \\
            TinyChart       & \xmark & 8.08 & 13.12 \\
            ChartMOE+PoT    & \xmark & 13.76 & 17.52 \\
            \bottomrule
        \end{tabular}%
    }
\end{wraptable}

\noindent\textbf{Out-of-Context Responses.} A recurring issue in MLLMs is their inability to remain grounded in the input when perturbations are introduced. Despite receiving clear instructions, such as “answer based on the image”, models frequently generate hallucinated responses. For example, we observed that in scenarios where a chart image is shifted and the relevant answer becomes invisible, models typically exhibit one of three behaviors: (1) hallucinating plausible answers, (2) making arithmetic guesses, or (3) relying on prior knowledge learned during training (knowledge leakage).

To further evaluate the impact of VPs, we conducted experiments using \emph{blank input images} with the top perfomance models on the clean ChartQA dataset. The results, summarized in Table~\ref{tab:blank_input_results}, reveal behavior contrary to what is expected with clear images. Typically, all models perform better on augmented datasets with explicit instruction-following tasks. However, when given blank inputs, the models fail to follow instructions. For arithmetic guessing, an example is the question: “\textit{Find the missing data in the sequence 24, \_ , 32, 33, 42?}” Most models guessed values between 28-30, which are close to the gold answer (29) and are often considered correct due to the relaxed accuracy metric allowing a 5\% margin. For knowledge-based leakage, questions like “\textit{What is the major cause of death in the U.S.?}” are answered using prior knowledge from the training data. Despite the blank input, the models often correctly provide the ground truth answer (e.g., Heart disease), highlighting reliance on external knowledge rather than visual input \& instruction.

\noindent\textbf{Spatial Understanding.} MLLMs perform poorly in understanding complex charts, \ie, positional- and structural relationships, resulting in more than 10\% of the errors involving relational inference. For example, extracting metadata from charts can follow various approaches: beyond using x-y axis labels to estimate point values, plot textual annotations above the bars, data points or arrows directly specifying numbers can serve as alternative techniques. General- and document-related MLLMs demonstrate better spatial reasoning, due to their exposure to localization tasks during pretraining, such as visual grounding and spatial alignment. Models like LLaVA-OneVision, Qwen-VL, and UReader benefit from these pretraining strategies.

\noindent\textbf{Fine-tuning vs. Training from Scratch.} All chart-related models, which are fine-tuned from general-purpose MLLMs, exhibit weaker vision encoders and a higher number of visual hallucinations compared to those trained from scratch on chart analysis tasks. Thus, we can confirm the behavior of ``out-of-domain" performance degradation due to fine-tuning as highlighted by Niss \textit{et al.}~\cite{niss2024zero}. 
Furthermore, we observe that ``in-domain'' degradation becomes more pronounced with fine-tuning when a ``domain shift'' is present (\eg, scanned or captured charts). This issue may arise from the training strategy employed by chart-related models, which heavily rely on synthetic charts. This synthetic data often overlooks interpretative techniques while generating data with diverse styles, colors, and data types.

\subsection{Limitations}\label{sec4.4:limitations}

\noindent \textbf{Architectural Constraints.} Many existing chart-understanding models exhibit limited robustness due to reliance on simple fine-tuning techniques that leave the model architecture unchanged. These models often struggle to capture fine-grained visual cues or to ignore non-informative regions such as whitespace—both essential for accurate chart reasoning. Recent approaches like  MOEChart \cite{xu2024chartmoe} highlight the effectiveness of adapting intermediate layers or modular components addressing domain needs, leading to better alignment between model capacity and task complexity. 

\noindent \textbf{Synthetic Data.} Current models rely heavily on synthetic data, which -- despite offering structural diversity -- often do not capture the semantic and contextual cues present in real-world charts, leading to poor generalization. This issue could be mitigated by directing more effort toward the creation of realistic and real datasets that incorporate semantically rich charts. 

\noindent\textbf{Evaluation Metrics.} Another limitation we observed lies in the evaluation metric, namely relaxed accuracy. General-purpose and document-related models often generate longer and more comprehensive responses compared to those fine-tuned for chart-specific tasks, which tend to produce concise, single-word answers. The metric's reliance on string comparison may not accurately reflect the true performance of models in such cases. Additionally, the 5\% tolerance criterion presents challenges, particularly for numerical answers. For large values, such as 1,000, the allowable range ($\pm 50$) is significantly wider compared to smaller values, such as 1 ($\pm 0.05$). This discrepancy becomes problematic in year-based responses. 

\label{work_limitation}
\noindent\textbf{Work Limitations.} While CHAOS offers a robust evaluation framework for chart understanding, its applicability to other domains involving multimodal data remains limited and requires further investigation. The claims and insights presented in this work are grounded in our specific experimental setup and model selection, and thus should not be overgeneralized without additional cross-domain validation. Moreover, our human evaluation was conducted with 42 participants; expanding this study with a larger and more diverse user base could enhance the reliability of the defined difficulty levels and better reflect real-world variability in human interpretation.

\section{Conclusion}
In this work, we construct a comprehensive benchmark for \textbf{CH}art \textbf{A}nalysis with \textbf{O}utlier \textbf{S}amples (CHAOS), to assess the robustness of Multimodal Large Language Models (MLLMs) against real-world chart perturbations. There are 10 types of visual perturbations and 5 types of textural perturbations. Based on human evaluation, there are 3 severity levels defined for each perturbation. According to experiments on two chart analysis tasks, we obtain findings that reveal significant variability in MLLMs performance across different types and levels of perturbations, with chart-specific models often outperforming general-purpose and document-focused models on clean data but at the same time suffer more under perturbations. While MLLMs demonstrate resilience to minor visual and textual noise, severe perturbations introduce substantial challenges, frequently leading to hallucinations and misinterpretations. Our hallucination analysis and case studies provide further insights into the strengths and limitations of different MLLM architectures, reinforcing the importance of specialized chart-processing capabilities. We hope CHAOS will serve as a foundational benchmark to develop more robust MLLMs for real-world chart applications. 

\section{Acknowledgments}

This work was supported in part by Helmholtz Association of German Research Centers, in part by the Ministry of Science, Research and the Arts of Baden-Württemberg (MWK) through the Cooperative Graduate School Accessibility through AI-based Assistive Technology (KATE) under Grant BW6-03, and in part by Karlsruhe House of Young Scientists (KHYS). This work was partially performed on the HoreKa supercomputer funded by the MWK and by the Federal Ministry of Education and Research, partially on the HAICORE@KIT partition supported by the Helmholtz Association Initiative and Networking Fund, and partially on bwForCluster Helix supported by the state of Baden-Württemberg through bwHPC and the German Research Foundation (DFG) through grant INST 35/1597-1 FUGG.


{
\small
\bibliographystyle{plain}
\bibliography{main}
}

\clearpage

\appendix

\section{Human Evaluation Details}
\label{supply:human_eval_details}
\begin{figure}[hb]
    \centering
    \includegraphics[width=0.8\linewidth]{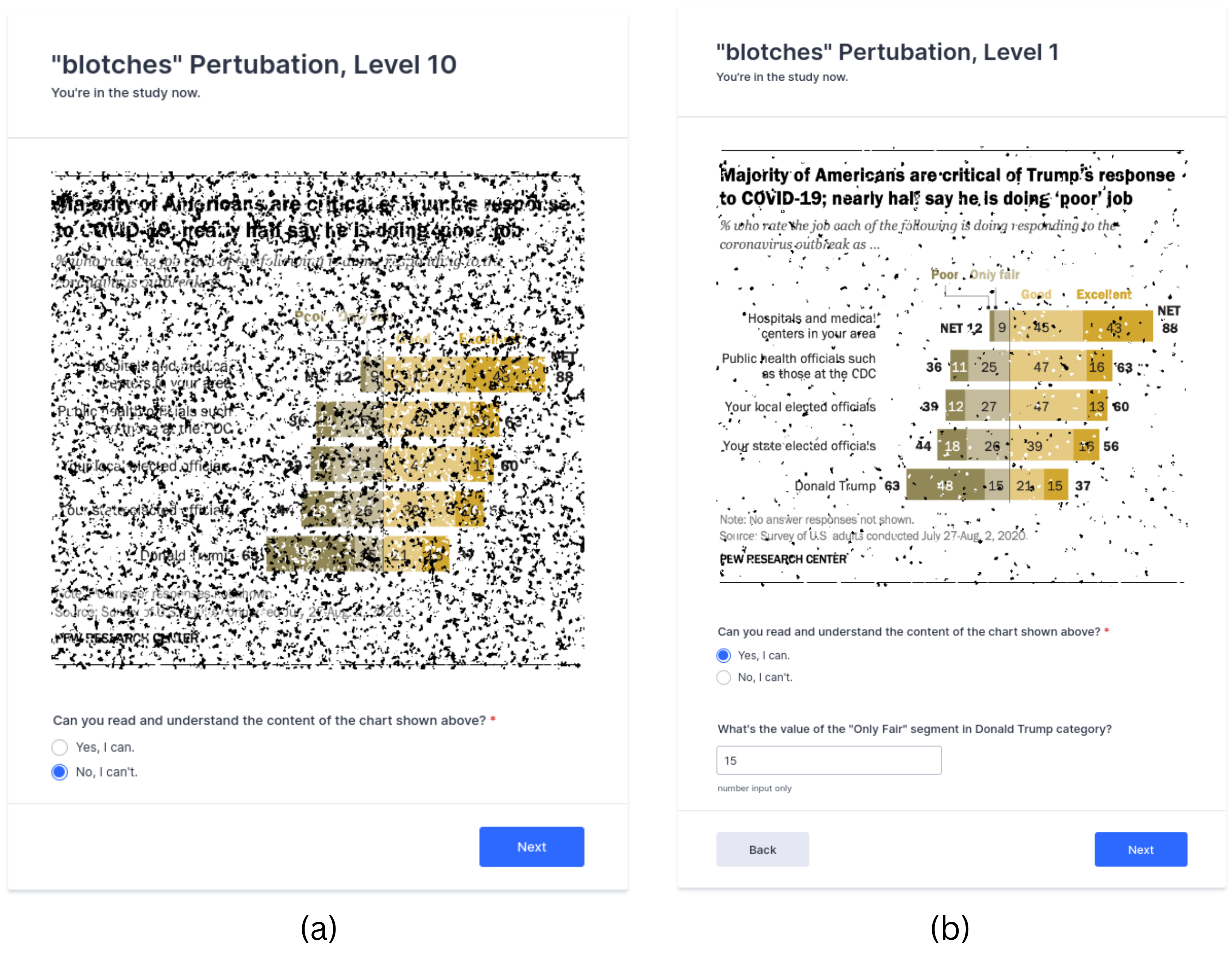}
    \caption{Study design. Participants start at (a) the highest perturbation level (Level 10) for each chart. (b) Upon confirming the level is interpretable, a corresponding question is posted.}
    \label{fig:userstudy}
\end{figure}

To conduct the user study, participants were invited to complete a pre-designed form hosted on the online platform JotForm. The form began with a trial perturbation to familiarize participants with the process. Subsequently, for each perturbation level, participants were instructed to indicate whether they could "see the image and answer a related question" or if the perturbation level needed to be reduced. Figure \ref{fig:userstudy} provides a screenshot of the user study form for the speckle perturbation case. To minimize bias from prior knowledge of the image content without noise, the study started with the highest perturbation level (level 10, maximum severity) instead of beginning with the clean image.

\begin{figure}[ht!]
    \centering
    \includegraphics[width=0.7\linewidth]{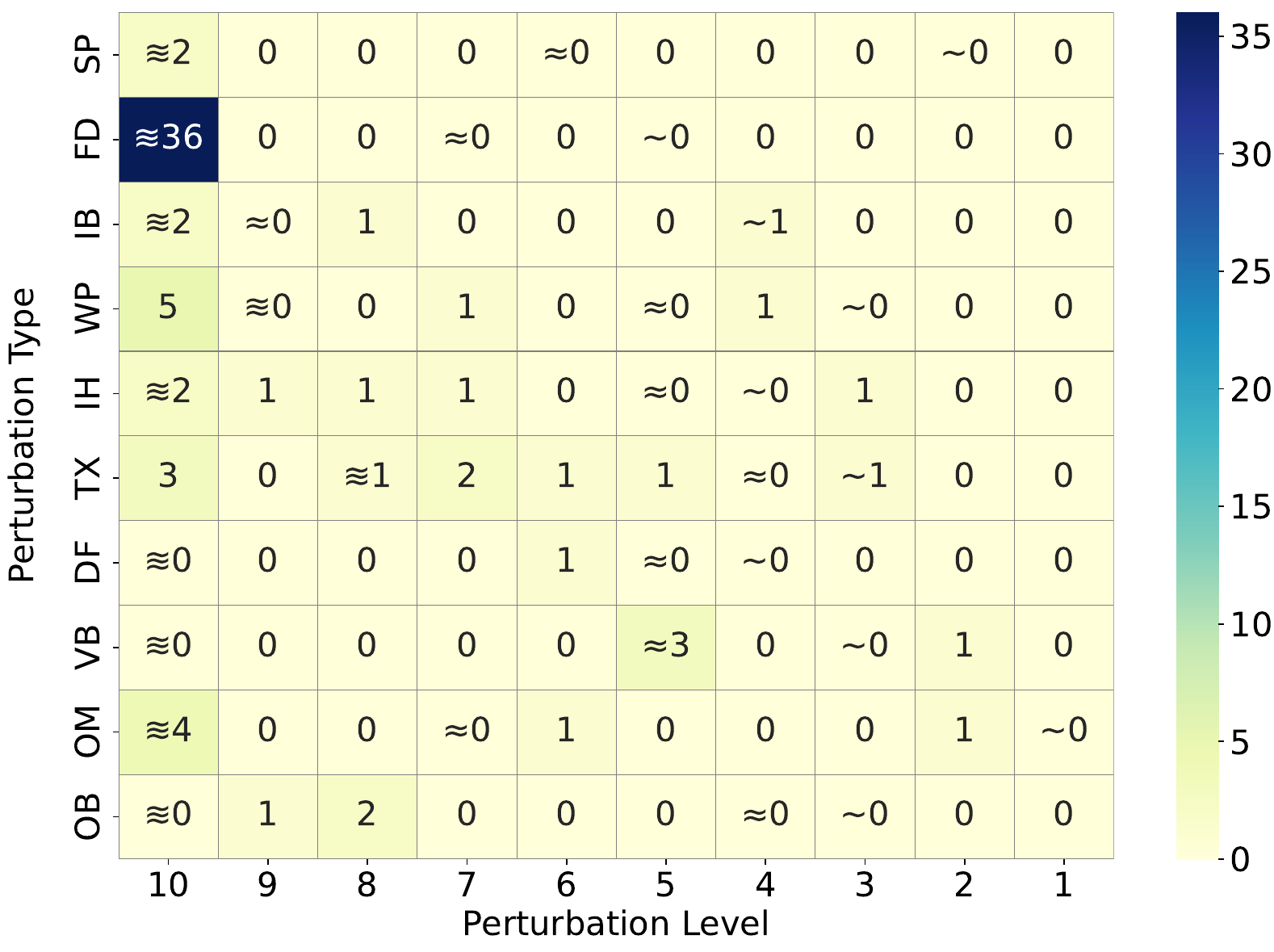}
    \caption{Distribution of human study results across perturbation types and levels. Each cell shows the number of participants who answered incorrect.
    \textit{Easy}, \textit{middle}, \textit{hard} levels are marked by symbols ($\sim$, $\approx$, $\equiv$) in the cell for each perturbation (each row).}
    \label{fig:incorrect}
\end{figure}

\begin{figure}[ht!]
    \centering
    \includegraphics[width=0.8\linewidth]{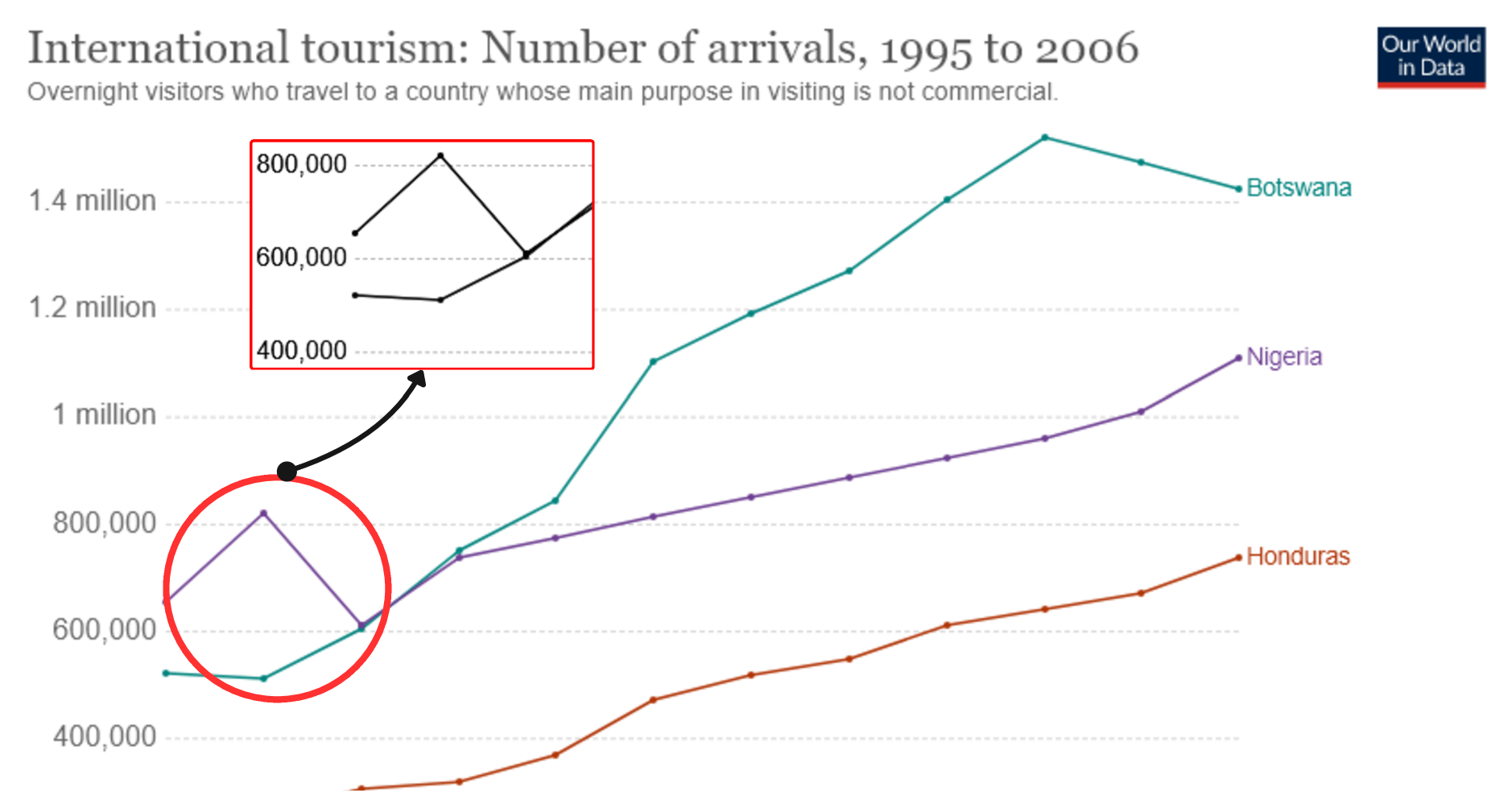}
    \caption{Chart with FD perturbation used in the user study. The zoomed-in bounding box highlights the level 10 FD perturbation case. The question presented was: "What is the approximate number of arrivals in Botswana in 1996?" and the GT: 500,000.}
    \label{fig:colorproblem}
\end{figure}

Additionally, Figure \ref{fig:incorrect} reports the frequency of incorrect responses, which was analyzed to identify potential outlier cases. One notable observation was the FD (fading) perturbation, where 75\% of participants were unable to provide correct answers. Figure 3 highlights the specific reason behind this difficulty. The FD perturbation significantly faded the image colors, which are essential for tracing and differentiating various chart lines. For instance, in Figure \ref{fig:colorproblem}, the overlapping green and purple lines caused confusion, making it challenging to determine which path corresponded to "Botswana." This tracing ambiguity is illustrated in the bounding box shown in Figure \ref{fig:colorproblem}.

\section{Implementation Details}

\subsection{Inference Setup}
\label{computer_resources}
\noindent\textbf{Hardware.} All models in this study were evaluated on one cluster node equipped with 4 A100 GPUs, each with 40 GB GPU memory.

\noindent\textbf{Prompts.} Table~\ref{tab:configuration_tokens_prompt} provides the prompts used for inference with each model. We adhered to the prompts reported in their original works to replicate the results accurately. Additionally, we maintained the same "maximum new tokens" length and ensured that the stop token was reached before exceeding this limit to enable a fair comparison. For the chart-to-text task, both the prompt and token count were standardized. The prompt used was: "Create a brief summarization or extract key insights based on the chart image," with a maximum token length of 2048.

\begin{table*}[ht!]
\caption{Prompt details for each model in the ChartQA task, used for both visual and textual perturbations.}
\vspace{-8pt}
\label{tab:configuration_tokens_prompt}
\centering
\resizebox{0.7\linewidth}{!}{ 
\setlength{\tabcolsep}{2.pt}
\begin{tabular}{|l|l|}
\hline

\rowcolor[gray]{0.9}
\textbf{Model} & \textbf{Inference Prompt} \\ \hline

\rowcolor{lightblue}
\multicolumn{2}{|l|}{\textbf{\emph{General}}} \\ 

\rowcolor{lightblue}
LLaVA-OneVision~\cite{li2024llava} & 
\begin{tabular}{p{0.5\linewidth}} 
\textbf{System}: \lq Answer the question with a single word.\rq ; \\
\textbf{User}: \{question\} \\  
\end{tabular}
\\ \hline

\rowcolor{lightblue}
Qwen-VL~\cite{bai2023qwen} &~\textbf{User}: \{image tokens\}, \{question\}, \lq Answer:\rq \\ \hline

\rowcolor{lightgreen}
\multicolumn{2}{|l|}{\textbf{\emph{Document-related}}} \\ 

\rowcolor{lightgreen}
DocOwl1.5~\cite{hu2024mplugdocowl} &~\textbf{User}: \{question\} \\ \hline

\rowcolor{lightgreen}
UReader~\cite{ye2023ureader} &~\textbf{User}: \lq Human:\rq \{image tokens\}, \lq Human:\rq \{question\}, \lq AI:\rq\\ \hline

\rowcolor{lightyellow}
\multicolumn{2}{|l|}{\textbf{\emph{Chart-related}}} \\ 

\rowcolor{lightyellow}
ChartInstruct~\cite{masry2024chartinstruct} &  \begin{tabular}{p{0.5\linewidth}} 
\textbf{System}: \lq Provide the final answer without repetition: \rq ; \\
\textbf{User}: \{image tokens\}, \lq Question:\rq \{question\}. \\
\end{tabular} \\ \hline

\rowcolor{lightyellow}
ChartLlama~\cite{han2023chartllama} & ~\textbf{User}: \{image tokens\}, \{question\} \\ \hline

\rowcolor{lightyellow}
ChartAst~\cite{meng2024chartassisstant} & \begin{tabular}{p{0.6\linewidth}}
\textbf{System}: \lq Below is an instruction that describes a task. Write a response that appropriately completes the request.\\
Instruction: \\
Please answer my question based on the chart:\rq ;\\
\textbf{User}: \{question\}, 'Response:'
\end{tabular}

\\ \hline

\rowcolor{lightyellow}
TinyChart@768~\cite{zhang2024tinychart} & \begin{tabular}{p{0.6\linewidth}}
\textbf{System}: \lq Answer with detailed steps.\rq \\
\textbf{User}: \{question\}
\end{tabular}\\ \hline

\end{tabular}}
\end{table*}

\subsection{Selection of Baselines}
Models were selected based on state-of-the-art performance at the time of this benchmark. Two MLLM was chosen for each of the three relevant categories, with additional models included for chart analysis to compare different LLMs and vision encoders for deeper insights.

\noindent\textbf{TinyChart}~\cite{zhang2024tinychart}: represents the current SOTA in the chart-related category with only 3 billion parameters. It offers an efficient architecture designed for chart understanding, represented by a modified ViT \cite{dosovitskiy2020image} architecture integrated with a Visual Token Merging module \cite{bolya2022token}, which aggregates similar visual tokens to reduce the length of visual feature sequences. This design allows the model to efficiently encode high-resolution chart images. Furthermore, TinyChart incorporates a Program-of-Thoughts (PoT) learning strategy, enabling the model to generate Python programs for numerical computations. This approach significantly enhances the ability to answer questions by involving mathematical reasoning.

\noindent\textbf{ChartAst}~\cite{meng2024chartassisstant}: Like TinyChart, it builds on general-purpose LVLMs; more specifically, the model is built on a Swin-BART encoder-decoder architecture. It adopts a one-stage instruction-tuning approach, eliminating the need for isolated projector training. The model leverages visualization tools, such as Matplotlib, to create large-scale synthetic charts with diverse styles and synthesizes charts with randomized attributes (e.g., color and fonts) using LLMs.

\noindent\textbf{ChartLlama}~\cite{han2023chartllama}:
Being the first to apply LLaVA-1.5~\cite{liu2023llava1.5} to ChartQA tasks. It utilized 160K instruction data generated by GPT-4, achieving impressive performance. The model follows the approach of continued training through tailored instruction-tuning and relies primarily on synthetic data generation.

\noindent\textbf{ChartInstruct}~\cite{masry2024chartinstruct}: It is the only model whose dataset is entirely composed of real-world charts. The model employs a two-stage training pipeline. In the initial stage, the focus is on aligning visual and textual representations, during which only the projector is trainable. In the second stage, both the projector and the text decoder are fine-tuned, while the visual encoder remains frozen.

\noindent\textbf{ChartMOE}~\cite{xu2024chartmoe}: uses Mixture-of-Experts (MoE) architecture, replacing the traditional linear projector with multiple task-specific connectors. Each expert is trained on distinct alignment tasks (chart-to-table, chart-to-JSON, chart-to-code) using the 900K-sample ChartMoE-Align dataset. A three-stage training pipeline enables the model to bridge modality gaps and achieve state-of-the-art performance.

\noindent\textbf{DocOwl1.5}~\cite{hu2024mplugdocowl}: Employs a unified instruction-tuning strategy to handle various domains, including documents, webpages, tables, charts, and natural images. Its visual encoder primarily relies on a pre-trained CLIPViT. A key factor in its architecture is the H-Reducer module, which encodes visual features while preserving image layout information. This is achieved by merging horizontally adjacent patches through convolution.

\noindent\textbf{DocOwl2}~\cite{hu2024mplugdocowl2}: introduces a High-resolution DocCompressor module that compresses each high-resolution document image into 324 tokens, guided by low-resolution global visual features. This layout-aware compression enables efficient processing of multi-page documents with reduced computational resources. The model employs a three-stage training framework—Single-image Pretraining, Multi-image Continue-pretraining, and Multi-task Finetuning.

\noindent\textbf{UReader}~\cite{ye2023ureader}:
It is designed for OCR-free multimodal understanding, targeting tasks across various document understanding domains. It is the only model achieving state-of-the-art performance in chart understanding tasks without requiring additional downstream fine-tuning. To enhance visual text understanding, it incorporates auxiliary tasks, such as reading text directly from images. The model introduces a shape-adaptive cropping module that processes high-resolution images by dividing them into multiple local segments. Each segment is independently encoded using a frozen visual encoder and a trainable visual abstractor.

\noindent\textbf{Qwen-VL}~\cite{bai2023qwen}:
Utilizes ViT architecture as the vision encoder, initialized with pre-trained weights from OpenCLIP's ViT-bigG model. To bridge the gap between the visual and textual modalities, Qwen-VL incorporates a position-aware adapter within its architecture. A notable feature of Qwen-VL is its capability to handle visual grounding task. 

\noindent\textbf{Intern VL2}~\cite{wang2024internvl2}: It employs a "ViT-MLP-LLM" architecture, integrating vision transformers (InternViT), MLP projectors, and large language models (e.g., InternLM2, Qwen2.5). The model utilizes a progressive scaling strategy, training from smaller to larger models while refining data quality, enhancing multimodal alignment and performance.

\noindent\textbf{Janus-Pro}~\cite{chen2025januspro}: is designed to handle both image understanding and generation tasks. It employs a decoupled visual encoding strategy, utilizing separate pathways for comprehension and generation, which are processed through a shared auto-regressive transformer architecture.

\noindent\textbf{LLaVA-OneVision}~\cite{li2024llava}:
A general-purpose LMM designed to excel in single-image, multi-image, and video scenarios. It integrates a SigLIP \cite{zhai2023sigmoid} vision encoder with a Qwen2 language backbone, enabling text generation conditioned on one or multiple images. To handle high-resolution inputs, it processes images using the anyres-9 technique \cite{liu2024llava}.

\section{Details of Perturbation Taxonomy}
In this section, we precisely describe the perturbation taxonomy, which is divided into two parts: \textit{Visual Perturbations} and \textit{Textual Perturbations}. Each perturbation is designed to approximate the realistic challenges of applying chart analysis systems in the real world. We detail the implementation and the varying severity levels for each perturbation type.

\subsection{Details of Visual Perturbations}

Visual perturbations mimic common perturbations that emerge in chart images due to environmental factors or device limitations. We introduce ten types of visual perturbations, each with increasing levels of severity to thoroughly evaluate the robustness of models in human recognition standards.

\noindent \textbf{(VP1) Defocus (DF)} simulates the effect of an out-of-focus camera lens, which causes the image to appear blurry. This effect is particularly common in photographs taken with shallow depth-of-field or due to incorrect focusing. This is implemented by convolving the original image $I$ with a Gaussian kernel $G_{\sigma}$:

\begin{equation} 
I' = I * G_{\sigma}, 
\end{equation}
where $*$ denotes the convolution operation, and $G_{\sigma}$ is a Gaussian kernel with standard deviation $\sigma$. The standard deviation $\sigma$ controls the intensity of the blur, \ie, higher values of $\sigma$ result in a more pronounced effect.

\noindent \textbf{(VP2) Vibration (VB)} simulates motion blur caused by camera movement during image capture, resulting in streaks and loss of detail in the chart elements. A linear motion blur kernel $K_{\text{v}}$ of length $L$ and angle $\theta$ is used:
\begin{equation}
I' = I * K_{\text{v}}(L, \theta), 
\end{equation}
where kernel length is set as $L = L_i$, where $L_i$ is the length under different severity, the angle $\theta$ is randomly selected in ranges $[0^\circ, 360^\circ]$ to simulate various motion directions.

\noindent \textbf{(VP3) Warping (WP)} introduces geometric distortions to the chart image, \eg, stretching or bending, which can occur from lens aberrations or improper scanning. This perturbation is 
implemented by applying a spatial transformation to the image $I$ using a distortion function $T(x, y)$. Specifically, pixels are mapped from their original coordinates $(x, y)$ to new coordinates:
\begin{equation}
\begin{aligned}
       T(x, y) = \alpha \cdot G_{\sigma}(R(x, y)), 
\end{aligned}
\end{equation}
\begin{equation}
\begin{aligned}
     \left\{\begin{matrix}
    x' = x + T_x(x, y) \\ y' = y + T_y(x, y), 
    \end{matrix}\right.
\end{aligned}
\end{equation}
where $T(x,y)$ is a random non-linear field that introduces displacement in both the $x$ and $y$ directions for warping effects. $G_\sigma$ is Gaussian Kernel with standard deviation $\sigma$ and the displacement intensity is increased by increasing the severity level.

\noindent \textbf{(VP4) Omission (OM)} involves removing or covering parts of the image, leading to incomplete information. This simulates scenarios where objects block the view or parts of the image are cut off. This is implemented by random shifts and rotations to the image $I$:
\begin{equation}
\begin{aligned}
    I'(x, y) = I\left( R^{-1} \cdot \begin{pmatrix} x - c_x \\ y - c_y \end{pmatrix} + \begin{pmatrix} c_x \\ c_y \end{pmatrix} - \begin{pmatrix} t_x \\ t_y \end{pmatrix} \right),
\end{aligned}
\end{equation}

\begin{equation}
    R = \begin{pmatrix}
\cos\theta & -\sin\theta \\
\sin\theta & \cos\theta
\end{pmatrix},
\end{equation}
where $\begin{pmatrix}
     t_x \\ t_y 
\end{pmatrix}$ is the random shift vector, $(c_x, c_y)$ is the rotation center and $\theta$ is random rotation angle.

\noindent \textbf{(VP5) Ink-Bleeding (IB)} simulates the diffusion of ink beyond intended boundaries, causing characters and lines to blur together, akin to low-quality prints or scans. To create this effect, we apply a morphological erosion operation for ink-bleeding to the image using an elliptical structuring element. The kernel size $K_e$ determines the extent of erosion, the larger the kernel, the more pronounced the ink bleeding effect. The basic mathematical definition of the erosion operation $\ominus$ of an image $A$ by a structuring element $B$ is as followed:
\begin{equation}
    (A \ominus B)(x, y) = \min_{(b_x, b_y) \in B}\{A(x + b_x, y + b_y)\} .
\end{equation}
To enhance the quality of the image during the erosion process, we first upscale the image by a factor of ten in both the horizontal and vertical dimensions. This upscaling allows for finer detail preservation when the erosion is applied. After the erosion operation, we downscale the image back to its original size.

\noindent \textbf{(VP6) Ink-Holdout (IH)} refers to the phenomenon that ink inadequately sticks to the printing surface, resulting in faded or incomplete lines and characters. To simulate this effect, we apply the morphological dilation operation, which is the mathematical counterpart to erosion. We apply the dilation operation with the same parameters in the erosion operation to ensure consistency simulating opposing behaviors. The dilation of the document image $A$ by an elliptical structuring element $B$ is mathematically defined as:

\begin{equation} (A \oplus B)(x, y) = \max_{(b_x, b_y) \in B} { A(x - b_x, y - b_y) }. \end{equation}

This operation effectively expands the lighter regions of the image, emulating areas where ink has insufficiently covered the substrate, producing the Ink-Holdout effect.

\noindent \textbf{(VP7) Obstacle (OB)} introduces shadow and glare that partially obstruct the chart. We introduce non-uniform illumination into document images for this perturbation. Mask $M$ is created with random polygons filled with black on a white canvas, which is then blurred using a Gaussian filter. The illumination adjustment can be described mathematically as a pixel-wise multiplication of the image $I$ with mask $M$ :
\begin{equation}
    I'(x, y) = V \cdot I(x, y) \cdot M(x, y) ,
\end{equation}
where $V$ is the illumination scaling factor, determined by the severity levels and type of illumination adjustment, \ie, shadow with $V_s$ and glare with $V_l$. 

\noindent \textbf{(VP8) Fading (FD)} models the gradual loss of color contrast, mimicking the effects of aging or 
exposure to harsh conditions. This effect is implemented by adjusting the brightness and contrast of the image $I$ using a linear transformation:
\begin{equation}
    I' = \alpha I + \beta, 
\end{equation}
where $\alpha$ controls contrast reduction (with $\alpha < 1$) and $\beta$ adjusts brightness. The severity levels correspond to decreasing $\alpha$ and adjusting $\beta$ to simulate more significant fading.

\noindent \textbf{(VP9) Speckle (SP)} adds speckle noise in document images. We overlay random Gaussian-distributed blobs onto the original image $I$. These blobs represent both foreground (dark) and background (light) noise components. The blobs are generated by randomly placing points within the image domain and applying Gaussian smoothing, with their density, size, and roughness controlled by a blob density factor $D_b$ corresponding to different severity levels.
The modified image is computed as:

\begin{equation} I' = \min\left( \max\left(I,\ N_{\text{fg}} \right),\ 1 - N_{\text{bg}} \right), \end{equation}
where $N_{\text{fg}}$ and $N_{\text{bg}}$ are the intensity maps of the foreground and background blob noise, respectively.

By adjusting $D_b$, we modulate the spatial distribution and intensity of the blobs, thus simulating varying degrees of speckle noise severity for robustness evaluation.

\noindent \textbf{(VP10) Texture (TX)} simulate texture interference patterns characteristic of document images. we replicate the complex plant fiber structures found in historical archival papers. The random paths of the fibers are modeled using a stochastic process. The final fibrous texture is obtained by blending the generated fiber patterns with the original image as follows:

\begin{equation} 
I' = M \cdot I_{\text{ink}} + (1 - M) \cdot (1 - I_{\text{paper}}) \times 255, 
\end{equation}
where $M$ is a mask that determines the application of ink ($I_{\text{ink}}$) and paper ($I_{\text{paper}}$) textures. The spatial distribution of fibers follows a Gaussian distribution to reflect the inherent randomness in paper composition. By adjusting the fiber density according to different noise levels, we accurately represent varying degrees of document wear and texture interference.

\subsection{Details of Textural Perturbations}
To thoroughly evaluate the model's robustness in textual queries, we introduce five types of textual perturbations in this subsection, each designed to simulate common mistakes encountered in natural language. Each perturbation type is applied with three levels of severity to emulate increasing degrees of distortion.

\noindent \textbf{(TP1) Character Adding (CA)} simulates the insertion of extraneous characters into the textual query, reflecting typographical errors or noise from defective input devices such as keyboards or speech recognition systems. Random characters are inserted into words at random positions within the text. The inserted characters are selected uniformly from the set of all lowercase and uppercase letters (a–z, A–Z) and digits (0–9). Given a textual sequence  $S = [s_1, s_2, \ldots, s_N]$, we introduce $K$ additional characters $c_k$ at positions $p_k$ , where $k = 1, \ldots, K$. The perturbed sequence $ S' $ is constructed as:
\begin{equation}
S' = [s_1, \ldots, s_{p_1 - 1}, c_1, \ldots, c_K, s_{p_K}, \ldots, s_N],
\end{equation}
where $ K $ is determined by the severity level, and $ c_k $ are randomly selected characters.

\noindent \textbf{(TP2) Character Deletion (CD)} involves the accidental omission of characters from the text, which can alter word structures and potentially change the intended meaning. Characters are randomly deleted from words within the text. Deletion positions are chosen uniformly at random, excluding spaces and punctuation marks. Given $ S = [s_1, s_2, \ldots, s_N] $, we remove $ K $ characters at positions $ p_k $. The perturbed sequence $ S' $ is:
\begin{equation}
S' = S \setminus \{ s_{p_1}, s_{p_2}, \ldots, s_{p_K} \},
\end{equation}
where $ \setminus $ denotes the set difference operator, effectively deleting the specified characters.

\noindent \textbf{(TP3) Character Replacement (CR)} substitutes correct characters with incorrect ones, reflecting misspellings or OCR errors. Characters in the text are replaced with random characters. Replacement positions are selected uniformly, and the new characters are chosen from the same set as in Character Addition. For $ S = [s_1, s_2, \ldots, s_N] $, we replace $ K $ characters at positions $ p_k $ with new characters $ c_k $:
\begin{equation}
s'_{p_k} = c_k,\quad \text{for } k = 1, \ldots, K,
\end{equation}
where $ s'_{p_k} $ is the character at position $ p_k $ in the sequence $ S' $, and $ c_k $ are randomly selected replacement characters.

\noindent \textbf{(TP4) Character Swap (CS)} involves transposing adjacent characters within words, simulating ordinary typing errors such as transposition mistakes. Pairs of adjacent characters within words are swapped. Swap positions are selected randomly from words with a length of at least two characters. Given $ S = [s_1, s_2, \ldots, s_N] $, we perform $ K $ swaps at positions $ p_k $:
\begin{equation}
\begin{cases}
s'_{p_k} = s_{p_k + 1}, \\
s'_{p_k + 1} = s_{p_k},
\end{cases}\quad \text{for } k = 1, \ldots, K,
\end{equation}
where $ s'_{i} $ denotes the $ i $-th character in the perturbed sequence $ S' $.

\noindent \textbf{(TP5) Word Modification (WM)} mimics incorrect terms commonly found in daily language, which would appear due to misunderstandings or colloquial expressions. Words in the text are replaced with semantically or phonetically similar words. We utilize pre-trained word embeddings to find words that are close in semantic space or use a homonym dictionary for phonetically similar replacements.
Let $ W = [w_1, w_2, \ldots, w_M] $ be the sequence of words in the original text. We replace $ K $ words at positions $ q_k $ with modified words $ w'_{q_k} $:
\begin{equation}
w'_{q_k} = \operatorname{Modification}(w_{q_k}),\quad \text{for } k = 1, \ldots, K,
\end{equation}
where $ \operatorname{Modification} $ is a function mapping the original word $ w_{q_k} $ to a semantically similar word $ w'_{q_k} $.

\section{More Quantitative Results}

\subsection{Results of ChartQA}

We further analyze the effects of each perturbation on both the human and augmented splits, as shown in Tables~\ref{tab:visual_perturbations} and \ref{tab:textual_perturbations}. The ChartQA-human split, being based on human-generated questions, demands more arithmetic and calculation skills, as discussed in the main paper. In contrast, the augmented split is derived from template-based questions.

\noindent\circled{1} \textbf{Models exhibit lower robustness to speckle distortion.} The performance degradation caused by speckle perturbation (SP) is generally greater than that caused by other distortions, as shown in Table~\ref{tab:visual_perturbations}. For instance, ChartLlama demonstrates a significant drop in accuracy, from $\{60.08, 90.48\}$ on clean data to $\{27.28, 26.56\}$ under SP perturbation. 

\noindent\circled{2} \textbf{Fading perturbation had the least impact.} Among the tested perturbations, MLLM models demonstrated higher robustness to fading (FD) distortion compared to others. While color information can be critical for interpreting certain visualizations, as discussed in Figure~\ref{fig:colorproblem}, its significance was minimal in the ChartQA dataset. Specifically, only 2\% of the questions required accurate color information to arrive at the correct answer. 

\noindent\circled{3} \textbf{Augmented questions are generally easier to answer than human-generated questions under perturbations.} The template-based nature of augmented questions often focuses on locating specific values that are explicitly presented in the chart. This makes them less reliant on complex reasoning or interpretation, resulting in higher robustness to perturbations compared to human-generated questions, which typically require skills beyond OCR.
 
\begin{table*}[ht]
\caption{Detailed per-level relaxed accuracy results on the ChartQA dataset with \textbf{Visual Perturbations} at different difficulty levels (Easy, Medium, and Difficult). \textbf{Hum.}, \textbf{Aug.}, and \textbf{Avg.} represent human evaluation, augmented evaluation, and their average, respectively.}
\label{tab:visual_perturbations}
\resizebox{\textwidth}{!}{
\setlength{\tabcolsep}{2.pt}
\begin{tabular}{|l|cc|cc|cc|cc|cc|cc|cc|cc|cc|cc|cc|}
\hline

\rowcolor[gray]{0.9} 
\multicolumn{23}{|c|}{\textbf{Easy Level}} \\ \hline 

\textbf{Model} & \multicolumn{2}{c|}{\textbf{Clean}} & \multicolumn{2}{c|}{\textbf{SP}} & \multicolumn{2}{c|}{\textbf{FD}} & \multicolumn{2}{c|}{\textbf{IB}} & \multicolumn{2}{c|}{\textbf{WP}} & \multicolumn{2}{c|}{\textbf{IH}} & \multicolumn{2}{c|}{\textbf{TX}} & \multicolumn{2}{c|}{\textbf{DF}} & \multicolumn{2}{c|}{\textbf{VB}} & \multicolumn{2}{c|}{\textbf{OM}} & \multicolumn{2}{c|}{\textbf{OB}} \\ 
\cmidrule(lr){2-3} \cmidrule(lr){4-5} \cmidrule(lr){6-7} \cmidrule(lr){8-9} \cmidrule(lr){10-11} \cmidrule(lr){12-13} \cmidrule(lr){14-15} \cmidrule(lr){16-17} \cmidrule(lr){18-19} \cmidrule(lr){20-21} \cmidrule(l){22-23}

 & Hum. & Aug. & Hum. & Aug. & Hum. & Aug. & Hum. & Aug. & Hum. & Aug. & Hum. & Aug. & Hum. & Aug. & Hum. & Aug. & Hum. & Aug. & Hum. & Aug. & Hum. & Aug. \\ \hline

\rowcolor{lightblue}
\multicolumn{23}{|l|}{\textbf{\emph{General}}} \\ 

\rowcolor{lightblue} Intern-VL2~\cite{wang2024internvl2} & 75.28 & 94.88 & 62.88 & 86.40 & 73.28 & 94.40 & 74.88 & 94.48 & 68.32 & 88.48 & 73.04 & 94.08 & 65.04 & 80.72 & 73.76 & 93.76 & 72.08 & 93.52 & 72.40 & 91.84 & 72.48 & 94.00 \\

\rowcolor{lightblue} ChatGPT-4o~\cite{hurst2024gpt4o} & 74.00 & 70.96 & 63.36 & 61.52 & 73.52 & 71.28 & 73.60 & 70.24 & 70.24 & 67.36 & 71.20 & 70.16 & 71.12 & 69.12 & 72.80 & 70.48 & 71.76 & 68.64 & 71.20 & 69.44 & 71.44 & 69.04 \\ 

\rowcolor{lightblue}
LLaVA-OneVision~\cite{li2024llava} & 69.84 & 92.8 & 56.8 & 80.56 & 67.12 & 92.48 & 65.52 & 91.52 & 64.4 & 89.2 & 65.6 & 90.56 & 64.32 & 88.32 & 67.04 & 91.6 & 65.36 & 90.24 & 67.28 & 91.52 & 67.28 & 91.6 \\

\rowcolor{lightblue}
Qwen-VL~\cite{bai2023qwen} & 49.36 & 82.8 & 33.28 & 49.28 & 47.52 & 83.36 & 44.08 & 75.2 & 38.8 & 69.04 & 42.8 & 72.0 & 39.12 & 65.36 & 46.16 & 81.28 & 43.68 & 76.64 & 45.92 & 78.0 & 46.8 & 81.68 \\

\rowcolor{lightblue} Qwen2.5~\cite{Qwen2.5-VL} & 80.72 & 94.96 & 67.36 & 87.92 & 78.08 & 94.72 & 79.28 & 95.36 & 76.32 & 93.92 & 78.56 & 94.24 & 77.92 & 93.60 & 79.44 & 94.48 & 77.76 & 94.56 & 77.84 & 94.32 & 79.60 & 94.96 \\

\rowcolor{lightblue} Janus-Pro~\cite{chen2025januspro}  & 75.28 & 44.80 & 27.04 & 24.88 & 43.60 & 75.84 & 39.68 & 61.92 & 34.56 & 55.60 & 39.92 & 63.20 & 35.04 & 45.52 & 43.44 & 72.88 & 41.20 & 69.12 & 43.52 & 73.92 & 42.96 & 72.72 \\ \hline

\rowcolor{lightgreen}
\multicolumn{23}{|l|}{\textbf{\emph{Document-related}}} \\ 

\rowcolor{lightgreen}
DocOwl1.5~\cite{hu2024mplugdocowl} & 48.56 & 91.36 & 39.76 & 73.28 & 48.56 & 91.2 & 48.16 & 90.72 & 43.44 & 85.84 & 47.76 & 90.72 & 44.16 & 85.36 & 47.44 & 91.36 & 44.96 & 89.6 & 47.84 & 90.32 & 48.16 & 91.04 \\

\rowcolor{lightgreen}
UReader~\cite{ye2023ureader} & 39.28 & 79.12 & 32.32 & 54 & 37.92 & 76.72 & 37.44 & 73.6 & 34.08 & 65.76 & 36.08 & 70.56 & 33.76 & 56.48 & 39.52 & 75.44 & 35.36 & 65.44 & 38.8 & 76.48 & 39.28 & 78.64 \\

\rowcolor{lightgreen} DocOwl2.0~\cite{hu2024mplugdocowl2} & 47.60 & 91.76 & 38.88 & 76.88 & 47.52 & 91.12 & 46.80 & 90.64 & 43.20 & 87.28 & 46.08 & 90.24 & 42.48 & 87.28 & 46.96 & 90.48 & 44.48 & 89.44 & 47.76 & 89.92 & 46.64 & 91.36 \\ \hline

\rowcolor{lightyellow}
\multicolumn{23}{|l|}{\textbf{\emph{Chart-related}}} \\ 

\rowcolor{lightyellow}
ChartInstruct~\cite{masry2024chartinstruct} & 11.48 & 15.04 & 15.92 & 13.04 & 30.72 & 63.28 & 28.48 & 60.16 & 26.24 & 55.76 & 28.08 & 49.68 & 18.56 & 24.16 & 29.84 & 60.08 & 26.96 & 50.96 & 30.08 & 63.76 & 30.08 & 61.2 \\

\rowcolor{lightyellow}
ChartLlama~\cite{han2023chartllama} & 60.08 & 90.48 & 27.28 & 26.56 & 55.36 & 18.32 & 45.36 & 63.04 & 37.36 & 51.36 & 45.92 & 18 & 35.12 & 38.32 & 54.8 & 83.76 & 50 & 74.24 & 45.6 & 67.44 & 54.24 & 18.48 \\

\rowcolor{lightyellow}
ChartAst~\cite{meng2024chartassisstant} &44.72 & 68.56 & 30.32 & 30.64 & 44.96 & 69.76 & 42.48 & 63.76 & 35.28 & 45.92 & 41.44 & 63.04 & 31.84 & 31.28 & 43.76 & 68.8 & 41.76 & 64.16 & 41.92 & 64.4 & 43.36 & 66.64 \\

\rowcolor{lightyellow}
TinyChart@768~\cite{zhang2024tinychart} & 57.92 & 94.8 & 44.48 & 74.56 & 58.16 & 94.24 & 56 & 92.24 & 51.44 & 90.64 & 54.72 & 90.24 & 53.44 & 92.24 & 56.4 & 93.84 & 48.16 & 84.56 & 57.76 & 93.76 & 54.48 & 93.28 \\

\rowcolor{lightyellow} ChartMOE-PoT~\cite{xu2024chartmoe} & 78.08 & 90.96 & 58.96 & 67.84 & 75.92 & 91.44 & 74.72 & 87.36 & 70.64 & 83.44 & 69.36 & 83.20 & 70.48 & 79.20 & 75.76 & 89.36 & 74.24 & 87.12 & 75.12 & 89.68 & 76.16 & 90.08 \\ \hline

\hline
\rowcolor[gray]{0.9} 
\multicolumn{23}{|c|}{\textbf{Medium Level}} \\ \hline 

\rowcolor{lightblue}
\multicolumn{23}{|l|}{\textbf{\emph{General}}} \\ 

\rowcolor{lightblue} Intern-VL2~\cite{wang2024internvl2} & 75.28 & 94.88 & 39.68 & 47.44 & 71.68 & 93.52 & 63.52 & 88.48 & 60.80 & 79.12 & 68.00 & 92.24 & 54.08 & 53.76 & 70.16 & 92.32 & 52.24 & 72.24 & 46.32 & 47.84 & 69.68 & 93.44 \\

\rowcolor{lightblue} ChatGPT-4o~\cite{hurst2024gpt4o} & 74.00 & 70.96 & 44.32 & 38.48 & 72.32 & 70.32 & 69.20 & 66.80 & 64.80 & 63.04 & 65.76 & 66.16 & 68.37 & 65.12 & 73.68 & 69.52 & 57.92 & 55.20 & 51.04 & 46.32 & 69.28 & 70.24 \\ 

\rowcolor{lightblue}
LLaVA-OneVision~\cite{li2024llava} & 69.84 & 92.8 & 39.12 & 50.8 & 66.16 & 92.16 & 57.04 & 81.92 & 59.12 & 84.08 & 61.2 & 87.12 & 61.84 & 85.04 & 65.52 & 90.88 & 46 & 61.28 & 44.56 & 52.16 & 66.72 & 91.28 \\

\rowcolor{lightblue}
Qwen-VL~\cite{bai2023qwen} & 49.36 & 82.8 & 20.56 & 14.8 & 46.88 & 83.36 & 32.32 & 42.56 & 30.64 & 51.28 & 40 & 64.24 & 33.12 & 44 & 44.72 & 76.72 & 30.96 & 35.68 & 30.4 & 36.72 & 44.72 & 80.96 \\

\rowcolor{lightblue} Qwen2.5~\cite{Qwen2.5-VL} & 80.72 & 94.96 & 51.28 & 64.24 & 76.80 & 94.72 & 74.56 & 88.40 & 69.84 & 89.68 & 72.64 & 93.20 & 76.32 & 92.00 & 78.24 & 94.08 & 54.00 & 65.84 & 47.60 & 48.80 & 78.16 & 94.32 \\

\rowcolor{lightblue} Janus-Pro~\cite{chen2025januspro}  & 75.28 & 44.80 & 18.00 &  9.20 & 42.08 & 76.16 & 25.52 & 19.76 & 29.76 & 39.52 & 35.52 & 58.48 & 30.08 & 26.64 & 42.64 & 69.52 & 30.48 & 42.88 & 31.36 & 37.84 & 41.20 & 71.28 \\ \hline

\rowcolor{lightgreen}
\multicolumn{23}{|l|}{\textbf{\emph{Document-related}}} \\ 

\rowcolor{lightgreen}
DocOwl1.5~\cite{hu2024mplugdocowl} & 48.56 & 91.36 & 23.84 & 30.4 & 48.72 & 91.44 & 42.64 & 73.92 & 38.64 & 74.72 & 43.04 & 86 & 40.08 & 76.48 & 45.92 & 90.32 & 29.2 & 43.04 & 32.8 & 45.68 & 47.2 & 89.76 \\

\rowcolor{lightgreen}
UReader~\cite{ye2023ureader}& 39.28 & 79.12 & 22.88 & 26.8 & 37.36 & 75.36 & 32.32 & 49.52 & 31.12 & 48.48 & 33.44 & 62.72 & 29.12 & 45.76 & 38.32 & 72.4 & 24.4 & 23.6 & 30 & 42.96 & 39.76 & 77.44 \\

\rowcolor{lightgreen} DocOwl2.0~\cite{hu2024mplugdocowl2} & 47.60 & 91.76 & 22.40 & 19.84 & 46.96 & 90.80 & 41.44 & 72.88 & 38.32 & 76.40 & 43.04 & 87.12 & 41.84 & 77.76 & 45.04 & 89.28 & 30.16 & 41.68 & 28.48 & 37.36 & 45.68 & 90.08 \\ \hline

\rowcolor{lightyellow}
\multicolumn{23}{|l|}{\textbf{\emph{Chart-related}}} \\ 

\rowcolor{lightyellow}
ChartInstruct~\cite{masry2024chartinstruct} & 11.48 & 15.04 & 11.44 & 4.96 & 30.48 & 63.52 & 15.92 & 11.84 & 22.56 & 39.44 & 25.2 & 37.84 & 14.72 & 8.8 & 26.64 & 55.68 & 17.28 & 16.24 & 22.72 & 34.24 & 27.92 & 60 \\

\rowcolor{lightyellow}
ChartLlama~\cite{han2023chartllama} & 60.08 & 90.48 & 19.92 & 16.96 & 53.52 & 18.24 & 31.52 & 29.44 & 31.6 & 38.72 & 42 & 17.6 & 29.12 & 26.64 & 51.76 & 76.64 & 36.88 & 51.36 & 32.08 & 38.96 & 51.52 & 18.24  \\

\rowcolor{lightyellow}
ChartAst~\cite{meng2024chartassisstant} & 44.72 & 68.56 & 16.56 & 8.0 & 45.52 & 67.92 & 34.56 & 49.84 & 29.84 & 22.48 & 40.08 & 59.84 & 28.32 & 13.68 & 43.28 & 68.16 & 33.12 & 49.52 & 27.28 & 15.84 & 40.48 & 64.96 \\

\rowcolor{lightyellow}
TinyChart@768~\cite{zhang2024tinychart} & 57.92 & 94.8 & 25.68 & 35.04 & 56.96 & 94.56 & 31.52 & 29.44 & 43.04 & 79.44 & 48.08 & 82.08 & 50.88 & 84.48 & 50.16 & 87.92 & 22.56 & 20.72 & 39.04 & 51.04 & 52.08 & 90.72 \\ 

\rowcolor{lightyellow} ChartMOE-PoT~\cite{xu2024chartmoe} & 78.08 & 90.96 & 31.68 & 27.60 & 74.72 & 91.12 & 54.88 & 57.84 & 60.48 & 72.56 & 58.96 & 74.64 & 61.12 & 63.20 & 72.64 & 86.32 & 53.52 & 60.16 & 51.76 & 51.36 & 74.40 & 88.48 \\ \hline

\rowcolor[gray]{0.9} 
\multicolumn{23}{|c|}{\textbf{Difficult Level}} \\ \hline 

\rowcolor{lightblue}
\multicolumn{23}{|l|}{\textbf{\emph{General}}} \\ 

\rowcolor{lightblue} Intern-VL2~\cite{wang2024internvl2} & 75.28 & 94.88 & 24.24 & 19.20 & 61.92 & 82.64 & 35.84 & 45.68 & 38.24 & 43.52 & 40.08 & 63.68 & 27.84 & 14.24 & 27.60 & 16.00 & 20.56 & 15.12 & 38.40 & 36.56 & 50.08 & 72.24 \\ 

\rowcolor{lightblue} ChatGPT-4o~\cite{hurst2024gpt4o}\cite{hurst2024gpt4o} & 74.00 & 70.96 & 30.40 & 27.36 & 65.84 & 68.40 & 52.72 & 49.76 & 50.48 & 48.32 & 51.60 & 45.52 & 38.00 & 28.40 & 50.16 & 44.32 & 34.48 & 29.04 & 46.16 & 39.44 & 53.84 & 56.00 \\ 

\rowcolor{lightblue}
LLaVA-OneVision~\cite{li2024llava} & 69.84 & 92.8 & 27.68 & 30.16 & 61.04 & 91.28 & 34.96 & 48.48 & 41.68 & 60.16 & 41.04 & 57.92 & 30.32 & 27.12 & 34.16 & 33.68 & 23.04 & 19.28 & 37.6 & 42.32 & 46.32 & 68.4 \\

\rowcolor{lightblue}
Qwen-VL~\cite{bai2023qwen} & 49.36 & 82.8 & 16.24 & 9.84 & 44.8 & 82.72 & 25.12 & 19.44 & 22.48 & 22.8 & 26.88 & 31.52 & 17.36 & 9.28 & 26.64 & 24.4 & 16.96 & 10.48 & 26.88 & 26.32 & 32.88 & 51.36 \\ 

\rowcolor{lightblue} Qwen2.5~\cite{Qwen2.5-VL} & 80.72 & 94.96 & 35.60 & 37.52 & 67.84 & 92.56 & 52.72 & 65.20 & 47.28 & 63.68 & 46.00 & 67.84 & 42.80 & 36.00 & 44.56 & 59.04 & 20.64 & 11.84 & 39.52 & 36.24 & 55.36 & 75.60 \\

\rowcolor{lightblue} Janus-Pro~\cite{chen2025januspro}  & 75.28 & 44.80 & 16.80 &  9.20 & 39.12 & 73.68 & 21.36 & 12.08 & 21.92 & 20.96 & 26.88 & 33.28 & 17.84 &  9.28 & 28.32 & 27.68 & 18.48 & 11.44 & 25.68 & 32.80 & 26.80 & 38.72 \\ \hline

\rowcolor{lightgreen}
\multicolumn{23}{|l|}{\textbf{\emph{Document-related}}} \\ 

\rowcolor{lightgreen}
DocOwl1.5~\cite{hu2024mplugdocowl} & 48.56 & 91.36 & 20 & 14.16 & 48 & 91.2 & 26.56 & 37.04 & 27.44 & 32.56 & 26.48 & 36.88 & 24.16 & 15.76 & 19.52 & 10.48 & 17.44 & 10.56 & 28.56 & 35.52 & 37.76 & 67.36 \\

\rowcolor{lightgreen}
UReader~\cite{ye2023ureader} & 39.28 & 79.12 & 17.84 & 12.72 & 36 & 72.88 & 23.2 & 20.8 & 22.8 & 23.28 & 21.6 & 22.4 & 20.24 & 13.04 & 21.44 & 17.36 & 17.84 & 12 & 27.04 & 36.24 & 32.4 & 54.8 \\

\rowcolor{lightgreen} DocOwl2.0~\cite{hu2024mplugdocowl2} & 47.60 & 91.76 & 18.40 &  8.00 & 45.76 & 90.24 & 25.92 & 36.88 & 27.20 & 33.52 & 24.48 & 37.44 & 22.48 & 11.04 & 18.16 &  9.44 & 16.80 & 10.32 & 24.48 & 27.20 & 36.32 & 69.60 \\ \hline

\rowcolor{lightyellow}
\multicolumn{23}{|l|}{\textbf{\emph{Chart-related}}} \\ 

\rowcolor{lightyellow}
ChartInstruct~\cite{masry2024chartinstruct} & 11.48 & 15.04 & 11.52 & 5.04 & 30.72 & 63.04 & 12.96 & 6.64 & 14.64 & 17.44 & 13.12 & 7.92 & 12 & 4.72 & 12.32 & 5.84 & 12.16 & 5.92 & 19.68 & 23.92 & 20.8 & 32.4 \\

\rowcolor{lightyellow}
ChartLlama~\cite{han2023chartllama} & 60.08 & 90.48 & 18.64 & 16.48 & 48.8 & 77.52 & 26.24 & 21.36 & 24.64 & 25.92 & 27.36 & 36.08 & 20.56 & 15.84 & 28.48 & 32.88 & 23.68 & 21.68 & 28 & 31.76 & 31.44 & 43.04  \\

\rowcolor{lightyellow}
ChartAst~\cite{meng2024chartassisstant} & 44.72 & 68.56 & 11.28 & 6.64 & 40.16 & 69.28 & 23.04 & 15.76 & 18.32 & 7.76 & 24.72 & 32.32 & 15.68 & 5.92 & 37.2 & 53.28 & 18.8 & 14.88 & 22.64 & 10.8 & 30.96 & 39.36 \\

\rowcolor{lightyellow}
TinyChart@768~\cite{zhang2024tinychart} & 57.92 & 94.8 & 18.08 & 15.76 & 54 & 93.6 & 21.76 & 12.72 & 28.16 & 43.92 & 22.96 & 22.16 & 20.72 & 10.16 & 18.16 & 11.76 & 17.04 & 9.92 & 33.6 & 41.04 & 30.08 & 51.2 \\

\rowcolor{lightyellow}  ChartMOE-PoT~\cite{xu2024chartmoe} & 78.08 & 90.96 & 23.52 & 14.96 & 67.76 & 89.44 & 35.04 & 24.88 & 38.08 & 41.28 & 36.72 & 41.36 & 27.12 & 15.36 & 37.28 & 35.68 & 27.36 & 20.48 & 45.68 & 41.28 & 49.92 & 64.56 \\ \hline

\end{tabular}}
\end{table*}

\definecolor{lightblue}{rgb}{0.85, 0.93, 1.0}
\definecolor{lightgreen}{rgb}{0.88, 1.0, 0.88}
\definecolor{lightyellow}{rgb}{1.0, 1.0, 0.85}

\begin{table*}[ht!]
\caption{Detailed per-level relaxed accuracy results on the ChartQA dataset with \textbf{Textual Perturbations}. Similar to table \ref{tab:visual_perturbations}.}
\label{tab:textual_perturbations}
\centering
\resizebox{1\textwidth}{!}{
\setlength{\tabcolsep}{10pt}
\begin{tabular}{|l|cc|cc|cc|cc|cc|cc|}
\hline

\rowcolor[gray]{0.9}
\multicolumn{13}{|c|}{\textbf{Easy Level}} \\ \hline 

\textbf{Model} & \multicolumn{2}{|c|}{\textbf{Clean}} & \multicolumn{2}{|c|}{\textbf{CA}} & \multicolumn{2}{|c|}{\textbf{CD}} & \multicolumn{2}{|c|}{\textbf{CR}} & \multicolumn{2}{|c|}{\textbf{CS}} & \multicolumn{2}{|c|}{\textbf{WM}} \\ \hline
 & Hum. & Aug. & Hum. & Aug. & Hum. & Aug. & Hum. & Aug. & Hum. & Aug. & Hum. & Aug. \\ \hline

\rowcolor{lightblue}
\multicolumn{13}{|l|}{\textbf{\emph{General}}} \\ 

\rowcolor{lightblue} Janus-Pro~\cite{chen2025januspro} & 75.28 & 44.80 & 38.88 & 72.24 & 38.96 & 68.80 & 35.52 & 65.68 & 38.48 & 68.72 & 33.12 & 62.16 \\ 

\rowcolor{lightblue} ChatGPT-4o~\cite{hurst2024gpt4o} & 74.00 & 70.96 & 71.68 & 69.84 & 69.12 & 65.76 & 67.92 & 66.00 & 69.60 & 62.48 & 59.60 & 64.08 \\ 

\rowcolor{lightblue} Intern-VL2~\cite{wang2024internvl2} & 75.28 & 94.88 & 69.68 & 92.40 & 68.16 & 90.24 & 63.20 & 89.68 & 66.48 & 90.56 & 61.20 & 90.16 \\ 

\rowcolor{lightblue}
LLaVA-OneVision~\cite{li2024llava} & 69.84 & 92.8 & 67.6 & 91.12 & 64.32 & 88.88 & 62.56 & 88.16 & 64.64 & 89.28 & 56.32 & 86.96 \\

\rowcolor{lightblue}
Qwen-VL~\cite{bai2023qwen} & 49.36 & 82.80 & 44.4 & 80.32 & 42.72 & 76.72 & 40.88 & 76.24 & 44.72 & 78.24 & 39.28 & 76.72 \\

\rowcolor{lightblue} Qwen-VL2.5~\cite{Qwen2.5-VL} & 80.72 & 94.96 & 75.60 & 94.32 & 73.84 & 90.80 & 71.60 & 92.08 & 74.48 & 91.52 & 65.20 & 90.24 \\ 

\rowcolor{lightgreen}
\multicolumn{13}{|l|}{\textbf{\emph{Document-related}}} \\ 

\rowcolor{lightgreen} DocOwl2.0~\cite{hu2024mplugdocowl2} & 47.60 & 91.76 & 45.20 & 89.36 & 43.28 & 86.32 & 41.60 & 86.48 & 42.88 & 88.00 & 36.24 & 83.68 \\ 

\rowcolor{lightgreen}
DocOwl1.5~\cite{hu2024mplugdocowl} & 48.56 & 91.36 & 46.88 & 89.36 & 45.2 & 86.4 & 42.48 & 86.08 & 45.84 & 86.0 & 38.96 & 85.2 \\

\rowcolor{lightgreen}
UReader~\cite{ye2023ureader} & 39.28 & 79.12 & 35.12 & 76.88 & 34.4 & 74.56 & 32.88 & 74.8 & 35.28 & 72.56 & 33.68 & 73.04 \\ \hline

\rowcolor{lightyellow}
\multicolumn{13}{|l|}{\textbf{\emph{Chart-related}}} \\ 

\rowcolor{lightyellow}
ChartInstruct~\cite{masry2024chartinstruct} & 11.48 & 15.04 & 29.28 & 58.56 & 29.36 & 54.0 & 26.4 & 56.4 & 29.84 & 52.32 & 24.96 & 44.48 \\

\rowcolor{lightyellow}
ChartLlama~\cite{han2023chartllama} & 60.08 & 90.48 & 53.28 & 87.2 & 53.2 & 85.44 & 49.92 & 83.76 & 52.64 & 84.56 & 44.0 & 17.76 \\

\rowcolor{lightyellow}
ChartAst~\cite{meng2024chartassisstant} & 44.72 & 68.56 & 40.48 & 65.92 & 40.24 & 64.96 & 39.2 & 63.84 & 41.44 & 62.88 & 32.64 & 56.08 \\

\rowcolor{lightyellow}
TinyChart@768~\cite{zhang2024tinychart} & 57.92 & 94.8 & 50.24 & 89.12 & 49.12 & 86.64 & 46.0 & 85.12 & 50.0 & 84.64 & 45.76 & 86.32 \\

\rowcolor{lightyellow} ChartMOE-PoE~\cite{xu2024chartmoe} & 78.08 & 90.96 & 73.04 & 89.36 & 70.48 & 86.24 & 66.64 & 85.44 & 71.76 & 88.24 & 64.24 & 84.88 \\ \hline

\rowcolor[gray]{0.9}
\multicolumn{13}{|c|}{\textbf{Medium Level}} \\ \hline 

\rowcolor{lightblue}
\multicolumn{13}{|l|}{\textbf{\emph{General}}} \\ 

\rowcolor{lightblue} Janus-Pro~\cite{chen2025januspro} & 75.28 & 44.80 & 34.32 & 67.76 & 34.64 & 65.12 & 29.04 & 56.88 & 34.08 & 63.04 & 27.60 & 57.28 \\ 

\rowcolor{lightblue} ChatGPT-4o~\cite{hurst2024gpt4o} & 74.00 & 70.96 & 71.84 & 70.24 & 65.76 & 61.52 & 64.24 & 64.40 & 65.76 & 57.84 & 49.84 & 57.20 \\

\rowcolor{lightblue} Intern-VL2~\cite{wang2024internvl2} & 75.28 & 94.88 & 64.72 & 90.88 & 60.88 & 87.28 & 53.36 & 84.72 & 59.28 & 86.00 & 53.76 & 84.40 \\

\rowcolor{lightblue}
LLaVA-OneVision~\cite{li2024llava} & 69.84 & 92.8 & 65.44 & 89.76 & 60.88 & 86.0 & 58.32 & 85.68 & 61.36 & 86.48 & 47.84 & 82.88 \\

\rowcolor{lightblue}
Qwen-VL~\cite{bai2023qwen} & 49.36 & 82.80 & 39.6 & 77.6 & 37.92 & 73.04 & 32.8 & 71.2 & 39.6 & 73.84 & 32.96 & 71.44 \\

\rowcolor{lightblue} Qwen-VL2.5~\cite{Qwen2.5-VL} & 80.72 & 94.96 & 72.48 & 93.20 & 68.88 & 87.52 & 63.92 & 86.72 & 69.68 & 86.88 & 56.16 & 86.40 \\ 

\rowcolor{lightgreen}
\multicolumn{13}{|l|}{\textbf{\emph{Document-related}}} \\ 

\rowcolor{lightgreen} DocOwl2.0~\cite{hu2024mplugdocowl2} & 47.60 & 91.76 & 40.64 & 87.52 & 38.96 & 81.84 & 37.52 & 80.48 & 40.32 & 82.56 & 31.60 & 78.72 \\ 

\rowcolor{lightgreen}
DocOwl1.5~\cite{hu2024mplugdocowl} & 48.56 & 91.36 & 44.64 & 88.24 & 40.8 & 82.08 & 37.68 & 80.96 & 43.04 & 80.0 & 34.16 & 79.6 \\

\rowcolor{lightgreen}
UReader~\cite{ye2023ureader} & 39.28 & 79.12 & 32.4 & 72.4 & 31.52 & 69.52 & 26.64 & 67.6 & 32.4 & 67.36 & 28.08 & 67.52 \\ \hline

\rowcolor{lightyellow}
\multicolumn{13}{|l|}{\textbf{\emph{Chart-related}}} \\ 

\rowcolor{lightyellow}
ChartInstruct~\cite{masry2024chartinstruct} & 11.48 & 15.04 & 25.76 & 53.28 & 24.88 & 48.64 & 21.76 & 44.96 & 24.64 & 42.96 & 20.88 & 37.68 \\

\rowcolor{lightyellow}
ChartLlama~\cite{han2023chartllama} & 60.08 & 90.48 & 48.56 & 19.68 & 47.12 & 81.44 & 42.4 & 76.64 & 46.4 & 78.8 & 37.12 & 76.88 \\

\rowcolor{lightyellow}
ChartAst~\cite{meng2024chartassisstant} & 44.72 & 68.56 & 38.48 & 61.04 & 36.32 & 58.56 & 31.92 & 55.36 & 35.68 & 57.84 & 29.2 & 50.48 \\

\rowcolor{lightyellow}
TinyChart@768~\cite{zhang2024tinychart} & 57.92 & 94.8 & 45.04 & 81.36 & 41.92 & 76.4 & 34.64 & 74.32 & 38.56 & 70.64 & 40.4 & 78.88 \\

\rowcolor{lightyellow} ChartMOE-PoE~\cite{xu2024chartmoe} & 78.08 & 90.96 & 68.56 & 86.88 & 62.64 & 83.52 & 57.84 & 80.88 & 63.76 & 82.96 & 54.16 & 79.76 \\ \hline

\rowcolor[gray]{0.9}
\multicolumn{13}{|c|}{\textbf{Difficult Level}} \\ \hline 

\rowcolor{lightblue}
\multicolumn{13}{|l|}{\textbf{\emph{General}}} \\ 

\rowcolor{lightblue} Janus-Pro~\cite{chen2025januspro} & 75.28 & 44.80 & 31.36 & 64.32 & 30.64 & 58.48 & 25.60 & 50.56 & 32.96 & 58.64 & 25.60 & 53.28 \\ 

\rowcolor{lightblue} ChatGPT-4o~\cite{hurst2024gpt4o} & 74.00 & 70.96 & 69.92 & 69.84 & 63.52 & 59.76 & 63.36 & 62.96 & 64.72 & 56.32 & 47.44 & 56.54 \\

\rowcolor{lightblue} Intern-VL2~\cite{wang2024internvl2} & 75.28 & 94.88 & 61.20 & 90.64 & 57.60 & 84.32 & 47.60 & 79.52 & 55.44 & 83.84 & 48.08 & 82.72 \\ 

\rowcolor{lightblue}
LLaVA-OneVision~\cite{li2024llava} & 69.84 & 92.8 & 64.0 & 90.4 & 57.84 & 84.16 & 55.12 & 81.76 & 60.88 & 85.04 & 43.36 & 79.6 \\

\rowcolor{lightblue}
Qwen-VL~\cite{bai2023qwen} & 49.36 & 82.80 & 38.4 & 75.44 & 37.84 & 70.4 & 30.64 & 66.64 & 37.12 & 71.68 & 29.68 & 70.32 \\

\rowcolor{lightblue} Qwen-VL2.5~\cite{Qwen2.5-VL} & 80.72 & 94.96 & 72.32 & 92.40 & 66.48 & 85.20 & 61.76 & 84.96 & 68.56 & 84.64 & 52.00 & 84.72 \\ 

\rowcolor{lightgreen}
\multicolumn{13}{|l|}{\textbf{\emph{Document-related}}} \\ 

\rowcolor{lightgreen} DocOwl2.0~\cite{hu2024mplugdocowl2}~\cite{hu2024mplugdocowl2} & 47.60 & 91.76 & 39.28 & 86.96 & 37.84 & 78.24 & 33.28 & 76.00 & 38.16 & 81.84 & 29.20 & 76.96 \\ 

\rowcolor{lightgreen}
DocOwl1.5~\cite{hu2024mplugdocowl} & 48.56 & 91.36 & 41.04 & 88.8 & 37.2 & 78.64 & 34.0 & 77.52 & 40.0 & 76.64 & 31.28 & 79.52 \\

\rowcolor{lightgreen}
UReader~\cite{ye2023ureader} & 39.28 & 79.12 & 29.12 & 71.68 & 30.0 & 66.08 & 22.24 & 62.88 & 30.8 & 63.6 & 25.44 & 66.64 \\ \hline

\rowcolor{lightyellow}
\multicolumn{13}{|l|}{\textbf{\emph{Chart-related}}} \\ 

\rowcolor{lightyellow}
ChartInstruct~\cite{masry2024chartinstruct} & 11.48 & 15.04 & 22.4 & 49.52 & 23.76 & 41.12 & 17.6 & 34.96 & 22.16 & 37.6 & 19.6 & 36.24 \\

\rowcolor{lightyellow}
ChartLlama~\cite{han2023chartllama} & 60.08 & 90.48 & 46.64 & 80.88 & 44.32 & 17.52 & 36.32 & 70.0 & 44.4 & 73.04 & 36.0 & 74.32 \\

\rowcolor{lightyellow}
ChartAst~\cite{meng2024chartassisstant} & 44.72 & 68.56 & 35.92 & 60.96 & 33.28 & 56.96 & 27.92 & 51.12 & 34.08 & 53.52 & 24.88 & 49.36 \\

\rowcolor{lightyellow}
TinyChart@768~\cite{zhang2024tinychart} & 57.92 & 94.8 & 37.2 & 75.28 & 35.2 & 67.76 & 30.4 & 56.88 & 32.48 & 63.84 & 34.32 & 75.68 \\

\rowcolor{lightyellow} ChartMOE-PoE~\cite{xu2024chartmoe} & 78.08 & 90.96 & 66.00 & 86.80 & 59.28 & 79.68 & 51.60 & 76.32 & 61.36 & 79.44 & 51.84 & 78.32 \\ \hline

\end{tabular}}
\end{table*}

\subsection{Results of Chart-to-Text}

The Chart-to-Text dataset consists of two chart sources: Pew and Statista. Statista provides visualizations on diverse topics such as politics, society, and health, while Pew focuses heavily on U.S. Politics \& Policy charts.

\noindent\circled{1} \textbf{Visual perturbations significantly impact chart-to-text summarization.} The summarization task demands careful attention to all pixels in the image, making it highly sensitive to distortions. MLLM models often struggle to maintain coherent output, even when only a small number of pixels are distorted at the easy perturbation level.

\noindent\circled{2} \textbf{Program-of-Thoughts improves the robustness of chart understanding model.} Employing techniques to learn interpretive strategies, such as Chain-of-Thoughts, as demonstrated by TinyChart~\cite{zhang2024tinychart}, results in higher robustness across all visual perturbations, as shown in Table~\ref{tab:textual_perturbations} and ~\ref{tab:visual_perturbations}.

\begin{table*}[ht!]
\caption{Detailed per-level BLEU-4 results on the Chart-to-text dataset with visual perturbations.}
\vspace{-8pt}
\label{tab:visual_perturbations_pew_Sta.}
\centering
\resizebox{\textwidth}{!}{
\setlength{\tabcolsep}{2.pt}
\begin{tabular}{|l|cc|cc|cc|cc|cc|cc|cc|cc|cc|cc|}
\hline

\rowcolor[gray]{0.9}
\multicolumn{21}{|c|}{\textbf{Easy Level}} \\ \hline

\textbf{Model} & \multicolumn{2}{c|}{\textbf{SP}} & \multicolumn{2}{c|}{\textbf{FD}} & \multicolumn{2}{c|}{\textbf{IB}} & \multicolumn{2}{c|}{\textbf{WP}} & \multicolumn{2}{c|}{\textbf{IH}} & \multicolumn{2}{c|}{\textbf{TX}} & \multicolumn{2}{c|}{\textbf{DF}} & \multicolumn{2}{c|}{\textbf{VB}} & \multicolumn{2}{c|}{\textbf{OM}} & \multicolumn{2}{c|}{\textbf{OB}} \\ 
\cmidrule(lr){2-3} \cmidrule(lr){4-5} \cmidrule(lr){6-7} \cmidrule(lr){8-9} \cmidrule(lr){10-11} \cmidrule(lr){12-13} \cmidrule(lr){14-15} \cmidrule(lr){16-17} \cmidrule(lr){18-19} \cmidrule(l){20-21}

& Pew & Sta. & Pew & Sta. & Pew & Sta. & Pew & Sta. & Pew & Sta. & Pew & Sta. & Pew & Sta. & Pew & Sta. & Pew & Sta. & Pew & Sta. \\ \hline

\rowcolor{lightyellow}
ChartInstruct~\cite{masry2024chartinstruct} & 4.97 & 1.86 & 6.52 & 5.7 & 6.34 & 4.07 & 5.87 & 4.23 & 5.85 & 4.55 & 5.85 & 1.79 & 6.25 & 5.14 & 5.47 & 4.74 & 6.39 & 5.47 & 6.53 & 5.56 \\

\rowcolor{lightyellow}
ChartLlama~\cite{han2023chartllama} & 4.01 & 0.49 & 5.66 & 1.69 & 5.56 & 1.26 & 4.59 & 0.98 & 5.44 & 1.92 & 4.98 & 0.87 & 5.72 & 1.57 & 5.41 & 1.49 & 5.52 & 1.49 & 5.78 & 1.47 \\

\rowcolor{lightyellow}
TinyChart@768~\cite{zhang2024tinychart} & 14.29 & 12.64 & 16.61 & 17 & 16.12 & 12.24 & 14.87 & 13.5 & 14.95 & 16.01 & 16.12 & 15.53 & 15.76 & 15.11 & 12.36 & 13.46 & 16.5 & 16.5 & 16.73 & 16.96 \\ \hline

\rowcolor[gray]{0.9}
\multicolumn{21}{|c|}{\textbf{Medium Level}} \\ \hline

\rowcolor{lightyellow}
ChartInstruct~\cite{masry2024chartinstruct} & 1.95 & 0.23 & 6.53 & 5.66 & 4.94 & 1.13 & 4.6 & 3.06 & 4.85 & 4 & 5.26 & 0.94 & 5.34 & 3.9 & 2.89 & 1.98 & 4.93 & 3.66 & 6.42 & 5.51 \\

\rowcolor{lightyellow}
ChartLlama~\cite{han2023chartllama} & 2.09 & 0.23 & 5.61 & 1.69 & 4.8 & 0.57 & 3.87 & 0.84 & 5.28 & 1.95 & 4.85 & 0.51 & 5.41 & 1.45 & 4.08 & 1.27 & 3.65 & 1.03 & 5.54 & 1.39 \\

\rowcolor{lightyellow}
TinyChart@768~\cite{zhang2024tinychart} & 6.84 & 2.87 & 16.91 & 16.95 & 8.13 & 3.29 & 11.72 & 9.9 & 11.21 & 14.35 & 15.35 & 10.83 & 12.94 & 11.44 & 2.67 & 3.71 & 10.86 & 9.46 & 16.37 & 16.73 \\ \hline

\rowcolor[gray]{0.9}
\multicolumn{21}{|c|}{\textbf{Difficult Level}} \\ \hline

\rowcolor{lightyellow}
ChartInstruct~\cite{masry2024chartinstruct} & 0.72 & 0.16 & 6.51 & 5.63 & 3.97 & 0.82 & 2.32 & 1.62 & 1.4 & 1.45 & 1.44 & 0.1 & 1.64 & 1.05 & 0.68 & 0.97 & 4.06 & 2.91 & 4.78 & 3.67 \\

\rowcolor{lightyellow}
ChartLlama~\cite{han2023chartllama} & 0.99 & 0.32 & 5.78 & 1.69 & 3.49 & 0.29 & 2.05 & 0.55 & 2.54 & 1.16 & 3.12 & 0.22 & 3.57 & 0.85 & 1.46 & 0.49 & 2.91 & 0.87 & 4.58 & 0.9 \\

\rowcolor{lightyellow}
TinyChart@768~\cite{zhang2024tinychart} & 2.14 & 0.54 & 16.92 & 16.99 & 4.71 & 1.74 & 4.94 & 5.21 & 1.25 & 2.66 & 4.9 & 0.28 & 1.23 & 0.55 & 0.37 & 0.4 & 7.93 & 5.62 & 10.65 & 9.92 \\ \hline

\end{tabular}}
\end{table*}

\newpage
\section{Qualitative Case Study}\label{qualti}
\begin{figure*}[ht!]
    \centering
    \includegraphics[width=1\linewidth]{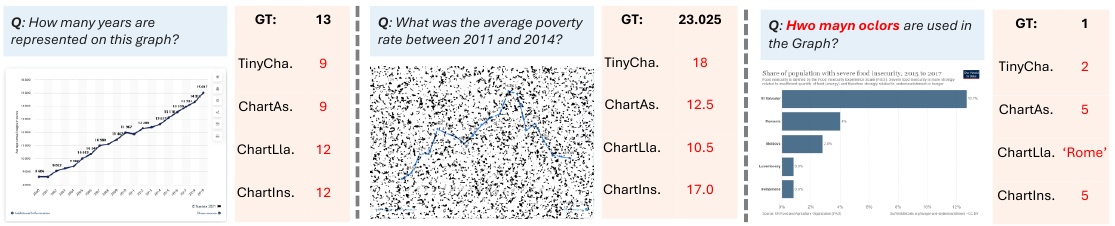}
    \begin{minipage}[t]{.33\textwidth}
        \vskip -3ex
        \subcaption{Case \circled{1} with VP: Ink-bleeding, \textit{easy}. }\label{fig_case_study_a}
    \end{minipage}%
    \begin{minipage}[t]{.33\textwidth}
        \vskip -3ex
        \subcaption{Case \circled{2} with VP: Speckle, \textit{hard}. }\label{fig_case_study_b}
    \end{minipage}%
    \begin{minipage}[t]{.33\textwidth}
        \vskip -3ex
        \subcaption{Case \circled{3} with TP: Chr Swap, \textit{middle}. }\label{fig_case_study_c}
    \end{minipage}%
    \vskip -1ex
    \caption{Case study of hallucinations across TP and VP. The samples are selected from cases where all models provided correct responses on clean inputs. Wrong answers by the models are marked in \textcolor{red}{red}, while the other models are correct. GT: Ground Truth. }
    \label{fig:case_study}
    \vskip -1ex
\end{figure*}

To delve deep into chart analysis, we conducted a qualitative analysis where models fail on perturbed chart images. Three cases are investigated as shown in Fig.~\ref{fig:case_study}.

\noindent \textbf{Case \circled{1}}: We identified 411 cases where general- and document-purpose models successfully responded under perturbations, but all chart-related models failed to produce the correct response. These cases, which we downsampled, were distributed across difficulty levels as follows: easy (153 cases), middle (143 cases), and hard (115 cases). Among these, nearly 90\% required arithmetic operations, exposing the limitations of vision encoders in handling arithmetic-based reasoning. In contrast, we observed only 80 cases where at least one chart-related model succeeded while others failed. The top sample in Fig.~\ref{fig_case_study_a} shows a line chart under an Ink-bleeding perturbation.

\noindent \textbf{Case \circled{2}}: Another example, illustrated by the middle sample in Fig.~\ref{fig_case_study_b}, underscores the interpretive capabilities of general-purpose models. Under severe speckle perturbations, the line chart required careful attention to locate the 2011 and 2014 labels and approximate the data point values from nearby points to calculate the average. This demonstrates the adaptability of general-purpose models in leveraging their vision encoders to infer missing or distorted information, an ability that chart-related models failed to develop in these scenarios.

\noindent \textbf{Case \circled{3}}: We identified 20 samples where text perturbations caused a significant drop in performance due to minimal character-level misspellings. Fig.~\ref{fig_case_study_c} highlights this sensitivity, even when a clean input image is provided. It demonstrates hallucinations across all chart-related models, triggered by swapping ``w" in ``How," ``y" in ``many," and ``c" in ``colors." This further underscores the vulnerability of text decoders when fine-tuned for chart-specific tasks.

To comprehensively introduce our findings on chart understanding model robustness, we arranged more detailed experimental results from the CHAOS benchmark.

\subsection{Sample Outputs}\label{qualti}

To further illustrate the robustness of chart analysis models, the following pages present examples from each perturbation level: easy (left), medium (center), and hard (right). Model responses are color-coded according to the legend provided below.

\begin{table}[H]
    \centering
    \caption{Color code legend for sample outputs.}
    \resizebox{0.6\columnwidth}{!}{%
        \begin{tabular}{|c|c|c|}
            \hline
            \cellcolor{blue!30}\textbf{General} & 
            \cellcolor{green!30}\textbf{Document-related} & 
            \cellcolor{orange!25}\textbf{Chart-related} \\ \hline
            \cellcolor{blue!15}Qwen-VL2.5~\cite{Qwen2.5-VL} & 
            \cellcolor{green!15}DocOwl2.0~\cite{hu2024mplugdocowl2} & 
            \cellcolor{orange!25}ChartMOE-PoT~\cite{xu2024chartmoe} \\ \hline
        \end{tabular}
    }
    \label{tab:legend}
\end{table}

\begin{table}[!htbp]
    \centering    
    \begin{tabular}{c}
        \includegraphics[width=\textwidth]{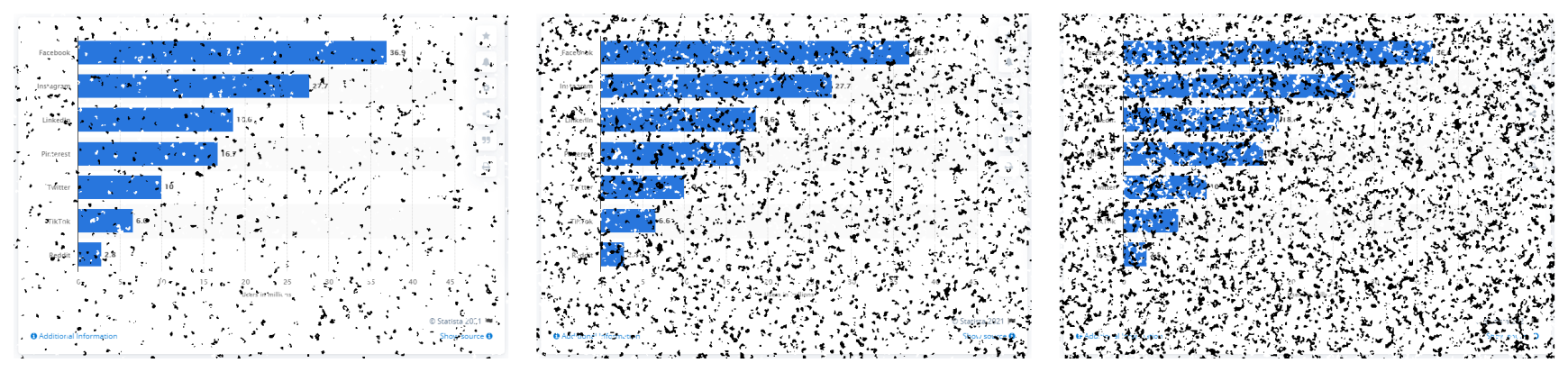} \\
    \end{tabular}
    \vspace{0.3cm}
    \begin{tabular}{p{\textwidth}}
        \centering
        \textbf{Perturb.:} SP $\circ$ 
        \textbf{Query:} Among Facebook, Instagram, and LinkedIn, what is the average minus the median? $\circ$ 
        \textbf{GT:} 3.38
    \end{tabular}
    \vspace{0.3cm}

    \resizebox{\textwidth}{!}{%
        \begin{tabular}{c@{\hspace{1.5cm}}c@{\hspace{1.5cm}}c}
            \begin{tabular}{|p{0.08\textwidth}|p{0.08\textwidth}|p{0.08\textwidth}|}
                \hline
                \cellcolor{blue!15}1.5 & \cellcolor{green!15}1.3 & \cellcolor{orange!25}10 \\ \hline
            \end{tabular}
            &
            \begin{tabular}{|p{0.08\textwidth}|p{0.08\textwidth}|p{0.08\textwidth}|}
                \hline
                \cellcolor{blue!15}10.5 & \cellcolor{green!15}11.5 & \cellcolor{orange!25}10.5 \\ \hline
            \end{tabular}
            &
            \begin{tabular}{|p{0.08\textwidth}|p{0.08\textwidth}|p{0.08\textwidth}|}
                \hline
                \cellcolor{blue!15}20 & \cellcolor{green!15}10 & \cellcolor{orange!25}0.15 \\ \hline
            \end{tabular}
        \end{tabular}
    }

    \label{tab:psample2}
\end{table}

\vspace{-0.2cm}

\begin{table}[!htbp]
    \centering    
    \begin{tabular}{c}
        \includegraphics[width=\textwidth]{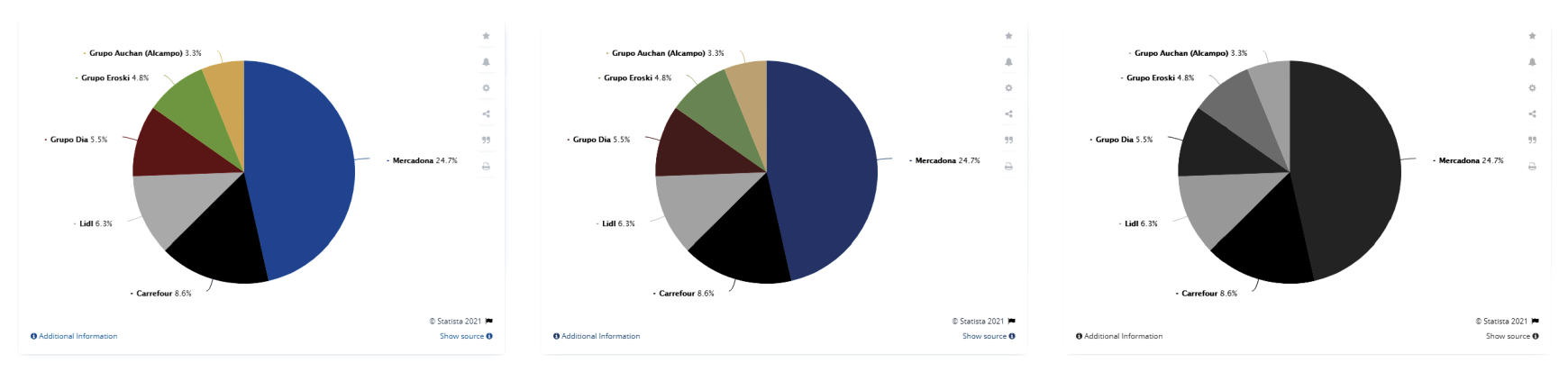} \\
    \end{tabular}
    
    \vspace{0.3cm}

    \begin{tabular}{p{\textwidth}}
        \centering
        \textbf{Perturb.:} FD $\circ$ 
        \textbf{Query:} What is the market share of Carrefour in Spain in 2020? $\circ$ 
        \textbf{GT:} 8.6
    \end{tabular}

    \vspace{0.3cm}

    \resizebox{\textwidth}{!}{%
        \begin{tabular}{c@{\hspace{1.5cm}}c@{\hspace{1.5cm}}c}
            \begin{tabular}{|p{0.08\textwidth}|p{0.08\textwidth}|p{0.08\textwidth}|}
                \hline
                \cellcolor{blue!15}8.5 & \cellcolor{green!15}8.6 & \cellcolor{orange!25}6.8 \\ \hline
            \end{tabular}
            &
            \begin{tabular}{|p{0.08\textwidth}|p{0.08\textwidth}|p{0.08\textwidth}|}
                \hline
                \cellcolor{blue!15}8.6 & \cellcolor{green!15}8.6 & \cellcolor{orange!25}8.6 \\ \hline
            \end{tabular}
            &
            \begin{tabular}{|p{0.08\textwidth}|p{0.08\textwidth}|p{0.08\textwidth}|}
                \hline
                \cellcolor{blue!15}8.6 & \cellcolor{green!15}8.6 & \cellcolor{orange!25}8.6 \\ \hline
            \end{tabular}
        \end{tabular}
    }

    \label{tab:psample3}
\end{table}

\vspace{-0.2cm}

\begin{table}[!htbp]
    \centering    
    \begin{tabular}{c}
        \includegraphics[width=\textwidth]{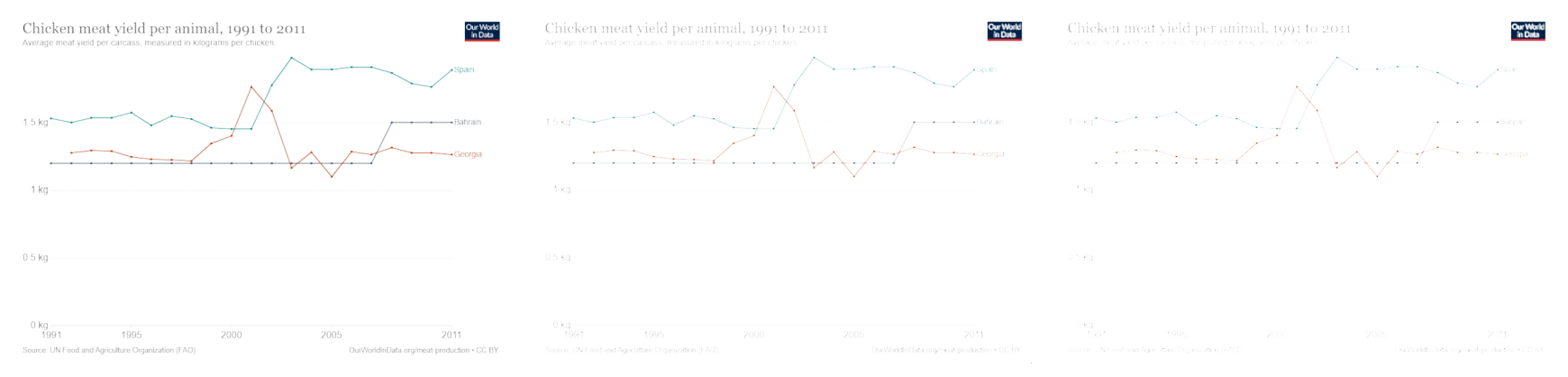} \\
    \end{tabular}
    
    \vspace{0.3cm}

    \begin{tabular}{p{\textwidth}}
        \centering
        \textbf{Perturb.:} IB $\circ$ 
        \textbf{Query:} Which country data is shown in the red line? $\circ$ 
        \textbf{GT Truth:} Georgia
    \end{tabular}

    \vspace{0.3cm}

    \resizebox{\textwidth}{!}{%
        \begin{tabular}{c@{\hspace{1.5cm}}c@{\hspace{1.5cm}}c}
            \begin{tabular}{|p{0.08\textwidth}|p{0.08\textwidth}|p{0.08\textwidth}|}
                \hline
                \cellcolor{blue!15}Georgia & \cellcolor{green!15}Spain & \cellcolor{orange!25}Bahrain \\ \hline
            \end{tabular}
            &
            \begin{tabular}{|p{0.08\textwidth}|p{0.08\textwidth}|p{0.08\textwidth}|}
                \hline
                \cellcolor{blue!15}Georgia & \cellcolor{green!15}Georgia & \cellcolor{orange!25}Bahrain \\ \hline
            \end{tabular}
            &
            \begin{tabular}{|p{0.08\textwidth}|p{0.08\textwidth}|p{0.08\textwidth}|}
                \hline
                \cellcolor{blue!25}Georgia & \cellcolor{green!15}Spain & \cellcolor{orange!25}Spain \\ \hline
            \end{tabular}
        \end{tabular}
    }

    \label{tab:psample1}
\end{table}

\vspace{-0.2cm}

\begin{table}[!htbp]
    \centering    
    \begin{tabular}{c}
        \includegraphics[width=\textwidth]{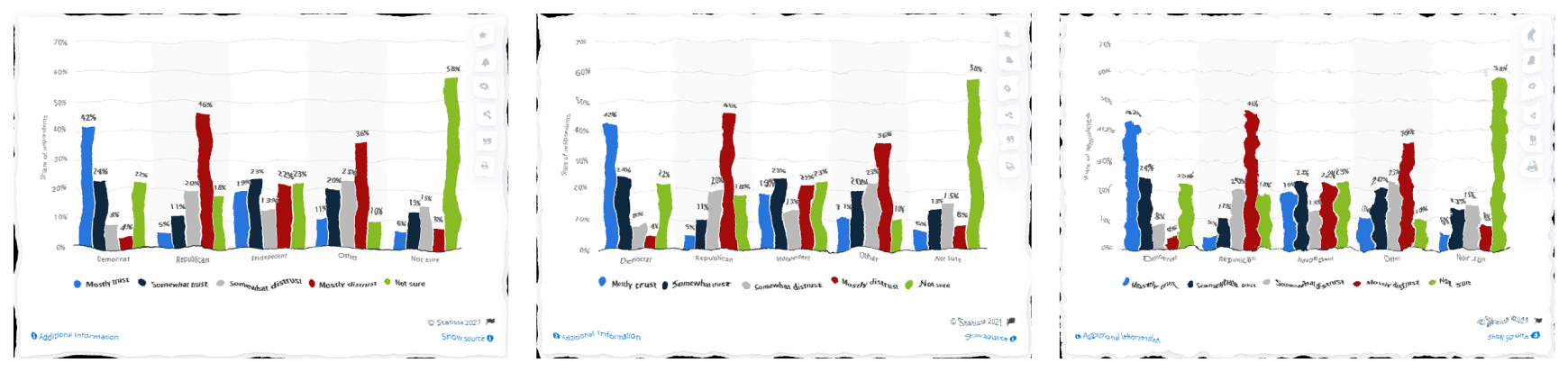} \\
    \end{tabular}
    
    \vspace{0.3cm}

    \begin{tabular}{p{\textwidth}}
        \centering
        \textbf{Perturb.:} WR $\circ$ 
        \textbf{Query:} What value is the tiniest bar? $\circ$ 
        \textbf{GT:} 4
    \end{tabular}

    \vspace{0.3cm}

    \resizebox{\textwidth}{!}{%
        \begin{tabular}{c@{\hspace{1.5cm}}c@{\hspace{1.5cm}}c}
            \begin{tabular}{|p{0.08\textwidth}|p{0.08\textwidth}|p{0.08\textwidth}|}
                \hline
                \cellcolor{blue!15}1 & \cellcolor{green!15}6 & \cellcolor{orange!25}5 \\ \hline
            \end{tabular}
            &
            \begin{tabular}{|p{0.08\textwidth}|p{0.08\textwidth}|p{0.08\textwidth}|}
                \hline
                \cellcolor{blue!15}1 & \cellcolor{green!15}6 & \cellcolor{orange!25}5 \\ \hline
            \end{tabular}
            &
            \begin{tabular}{|p{0.08\textwidth}|p{0.08\textwidth}|p{0.08\textwidth}|}
                \hline
                \cellcolor{blue!15}1 & \cellcolor{green!15}6 & \cellcolor{orange!25}3 \\ \hline
            \end{tabular}
        \end{tabular}
    }

    \label{tab:psample4}
\end{table}

\vspace{-0.2cm}

\begin{table}[!htbp]
    \centering    
    \begin{tabular}{c}
        \includegraphics[width=\textwidth]{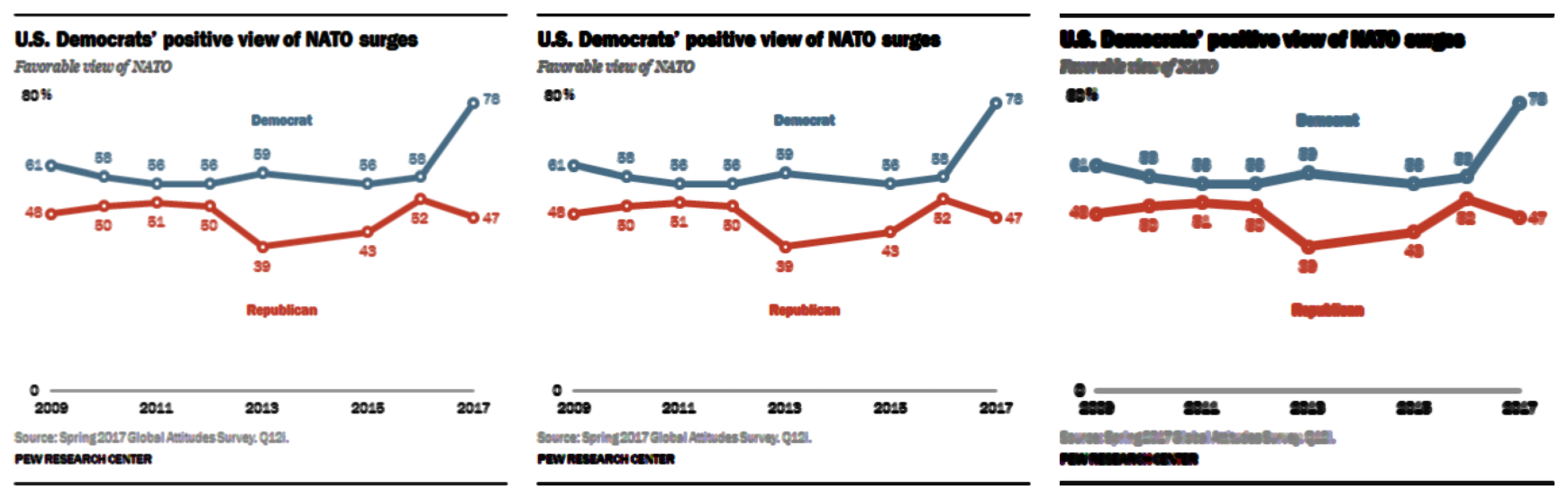} \\
    \end{tabular}
    
    \vspace{0.3cm}

    \begin{tabular}{p{\textwidth}}
        \centering
        \textbf{Perturb.:} IH $\circ$ 
        \textbf{Query:} How many points have 56 value in blue graph? $\circ$ 
        \textbf{GT:} 3
    \end{tabular}

    \vspace{0.3cm}

    \resizebox{\textwidth}{!}{%
        \begin{tabular}{c@{\hspace{1.5cm}}c@{\hspace{1.5cm}}c}
            \begin{tabular}{|p{0.08\textwidth}|p{0.08\textwidth}|p{0.08\textwidth}|}
                \hline
                \cellcolor{blue!15}2 & \cellcolor{green!15}2 & \cellcolor{orange!25}3 \\ \hline
            \end{tabular}
            &
            \begin{tabular}{|p{0.08\textwidth}|p{0.08\textwidth}|p{0.08\textwidth}|}
                \hline
                \cellcolor{blue!15}2 & \cellcolor{green!15}2 & \cellcolor{orange!25}3 \\ \hline
            \end{tabular}
            &
            \begin{tabular}{|p{0.08\textwidth}|p{0.08\textwidth}|p{0.08\textwidth}|}
                \hline
                \cellcolor{blue!15}2 & \cellcolor{green!15}2 & \cellcolor{orange!25}3 \\ \hline
            \end{tabular}
        \end{tabular}
    }

    \label{tab:psample5}
\end{table}

\vspace{-0.2cm}

\begin{table}[!htbp]
    \centering    
    \begin{tabular}{c}
        \includegraphics[width=\textwidth]{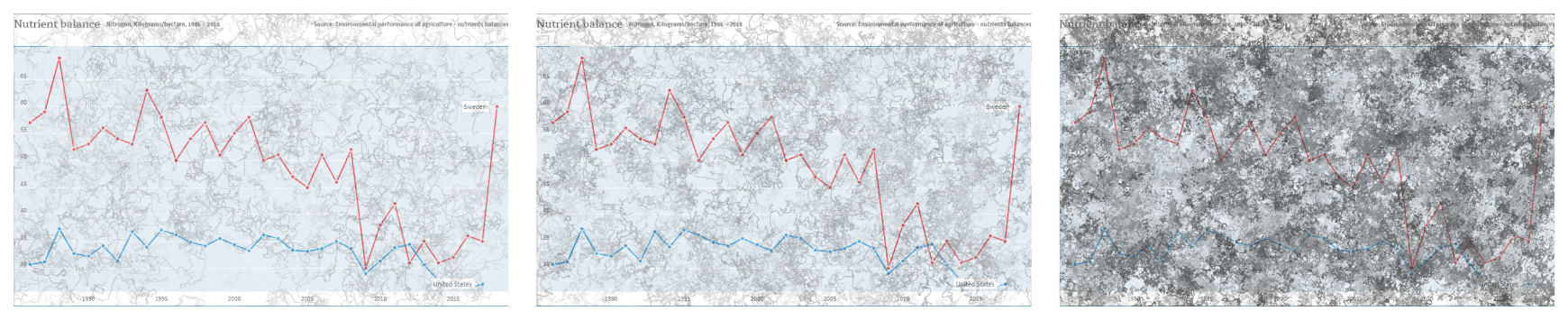} \\
    \end{tabular}
    
    \vspace{0.3cm}

    \begin{tabular}{p{\textwidth}}
        \centering
        \textbf{Perturb.:} TX $\circ$ 
        \textbf{Query:} What does the blue line refer to? $\circ$ 
        \textbf{GT:} US
    \end{tabular}

    \vspace{0.3cm}

    \resizebox{\textwidth}{!}{%
        \begin{tabular}{c@{\hspace{1.5cm}}c@{\hspace{1.5cm}}c}
            \begin{tabular}{|p{0.08\textwidth}|p{0.08\textwidth}|p{0.08\textwidth}|}
                \hline
                \cellcolor{blue!15}US & \cellcolor{green!15}Sweden & \cellcolor{orange!25}US \\ \hline
            \end{tabular}
            &
            \begin{tabular}{|p{0.08\textwidth}|p{0.08\textwidth}|p{0.08\textwidth}|}
                \hline
                \cellcolor{blue!15}US & \cellcolor{green!15}Sweden & \cellcolor{orange!25}US \\ \hline
            \end{tabular}
            &
            \begin{tabular}{|p{0.08\textwidth}|p{0.08\textwidth}|p{0.08\textwidth}|}
                \hline
                \cellcolor{blue!15}Sweden & \cellcolor{green!15}Sweden & \cellcolor{orange!25}Sweden \\ \hline
            \end{tabular}
        \end{tabular}
    }

    \label{tab:psample8}
\end{table}

\vspace{-0.2cm}

\begin{table}[!htbp]
    \centering    
    \begin{tabular}{c}
        \includegraphics[width=\textwidth]{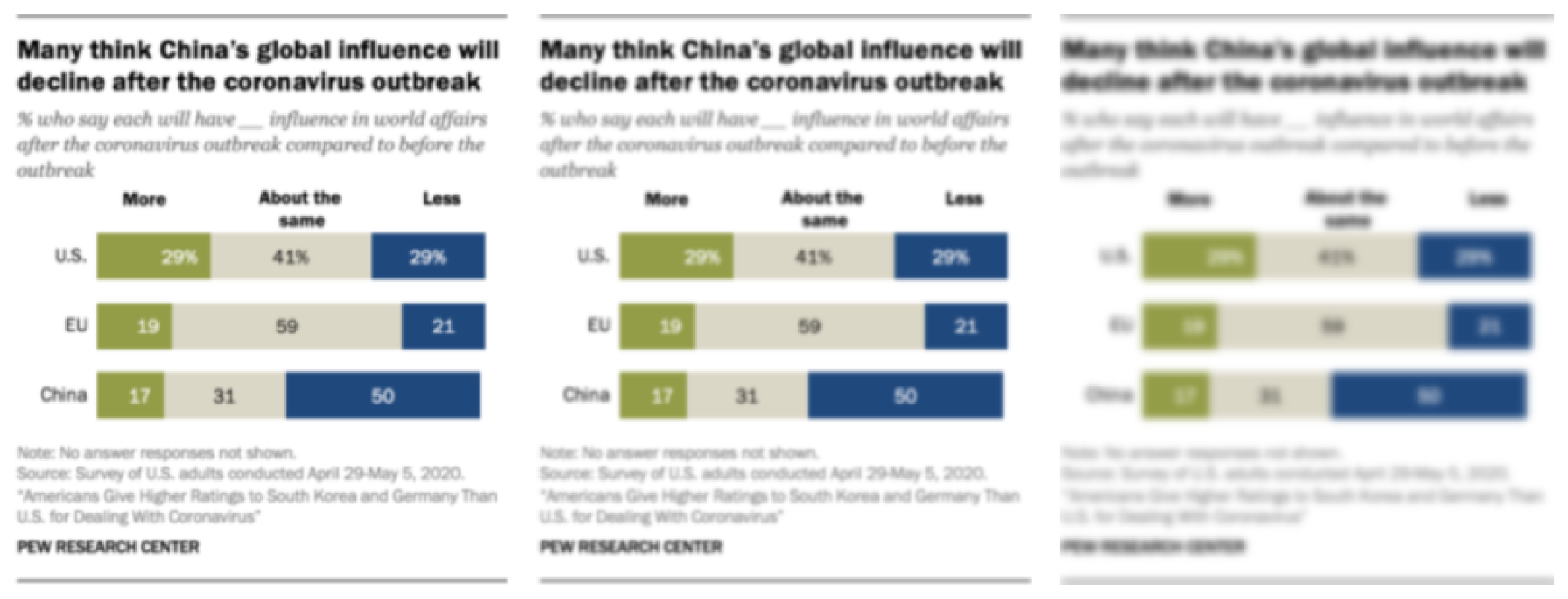} \\
    \end{tabular}
    
    \vspace{0.3cm}

    \begin{tabular}{p{\textwidth}}
        \centering
        \textbf{Perturb.:} DF $\circ$ 
        \textbf{Query:} What's the leftmost value of bar in China? $\circ$ 
        \textbf{GT:} 17
    \end{tabular}

    \vspace{0.3cm}

    \resizebox{\textwidth}{!}{%
        \begin{tabular}{c@{\hspace{1.5cm}}c@{\hspace{1.5cm}}c}
            \begin{tabular}{|p{0.08\textwidth}|p{0.08\textwidth}|p{0.08\textwidth}|}
                \hline
                \cellcolor{blue!15}17 & \cellcolor{green!15}1.7 & \cellcolor{orange!25}17 \\ \hline
            \end{tabular}
            &
            \begin{tabular}{|p{0.08\textwidth}|p{0.08\textwidth}|p{0.08\textwidth}|}
                \hline
                \cellcolor{blue!15}17 & \cellcolor{green!15}1.7 & \cellcolor{orange!25}17 \\ \hline
            \end{tabular}
            &
            \begin{tabular}{|p{0.08\textwidth}|p{0.08\textwidth}|p{0.08\textwidth}|}
                \hline
                \cellcolor{blue!15}17 & \cellcolor{green!15}1.7 & \cellcolor{orange!25}17 \\ \hline
            \end{tabular}
        \end{tabular}
    }

    \label{tab:psample6}
\end{table}

\vspace{-0.2cm}

\begin{table}[!htbp]
    \centering    
    \begin{tabular}{c}
        \includegraphics[width=\textwidth]{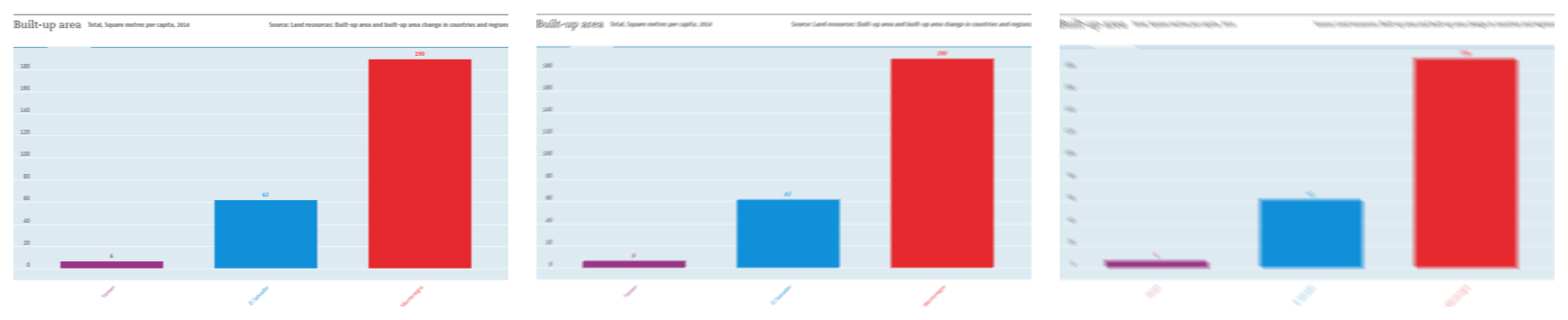} \\
    \end{tabular}
    
    \vspace{0.3cm}

    \begin{tabular}{p{\textwidth}}
        \centering
        \textbf{Perturb.:} VB $\circ$ 
        \textbf{Query:} What's the average value of Yemen and Montenegro? $\circ$ 
        \textbf{GT:} 98
    \end{tabular}

    \vspace{0.3cm}

    \resizebox{\textwidth}{!}{%
        \begin{tabular}{c@{\hspace{1.5cm}}c@{\hspace{1.5cm}}c}
            \begin{tabular}{|p{0.08\textwidth}|p{0.08\textwidth}|p{0.08\textwidth}|}
                \hline
                \cellcolor{blue!15}30 & \cellcolor{green!15}80 & \cellcolor{orange!25}110.5 \\ \hline
            \end{tabular}
            &
            \begin{tabular}{|p{0.08\textwidth}|p{0.08\textwidth}|p{0.08\textwidth}|}
                \hline
                \cellcolor{blue!15}30 & \cellcolor{green!15}80 & \cellcolor{orange!25}110.5 \\ \hline
            \end{tabular}
            &
            \begin{tabular}{|p{0.08\textwidth}|p{0.08\textwidth}|p{0.08\textwidth}|}
                \hline
                \cellcolor{blue!15}100 & \cellcolor{green!15}40 & \cellcolor{orange!25}50.5 \\ \hline
            \end{tabular}
        \end{tabular}
    }

    \label{tab:psample7}
\end{table}

\vspace{-0.2cm}

\begin{table}[!htbp]
    \centering    
    \begin{tabular}{c}
        \includegraphics[width=\textwidth]{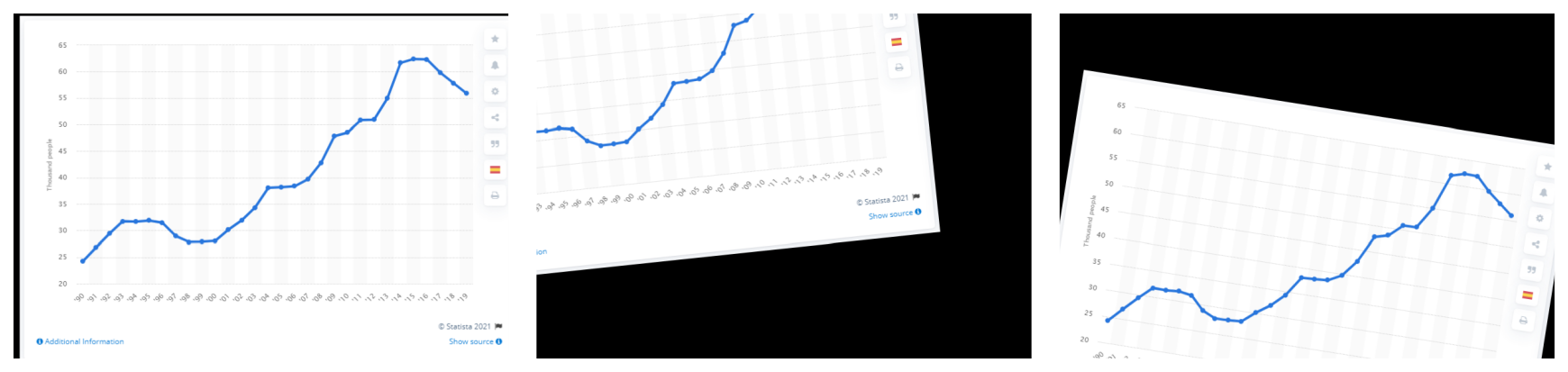} \\
    \end{tabular}
    
    \vspace{0.3cm}

    \begin{tabular}{p{\textwidth}}
        \centering
        \textbf{Perturb.:} OM $\circ$ 
        \textbf{Query:} How many Americans were covered by Medicaid in 2019? $\circ$
        \textbf{GT:} 55.85
    \end{tabular}

    \vspace{0.3cm}

    \resizebox{\textwidth}{!}{%
        \begin{tabular}{c@{\hspace{1.5cm}}c@{\hspace{1.5cm}}c}
            \begin{tabular}{|p{0.08\textwidth}|p{0.08\textwidth}|p{0.08\textwidth}|}
                \hline
                \cellcolor{blue!15}- & \cellcolor{green!15}33.8 & \cellcolor{orange!25}72.8 \\ \hline
            \end{tabular}
            &
            \begin{tabular}{|p{0.08\textwidth}|p{0.08\textwidth}|p{0.08\textwidth}|}
                \hline
                \cellcolor{blue!15}- & \cellcolor{green!15}33.8 & \cellcolor{orange!25}72.8 \\ \hline
            \end{tabular}
            &
            \begin{tabular}{|p{0.08\textwidth}|p{0.08\textwidth}|p{0.08\textwidth}|}
                \hline
                \cellcolor{blue!15}- & \cellcolor{green!15}50.3 & \cellcolor{orange!25}36.6 \\ \hline
            \end{tabular}
        \end{tabular}
    }

    \label{tab:psample9}
\end{table}

\end{document}